\RequirePackage{fix-cm}
\documentclass[twocolumn]{svjour3}          % twocolumn

\smartqed  % flush right qed marks, e.g. at end of proof
\usepackage{times}
\usepackage{epsfig}
\usepackage{graphicx}
\usepackage{amsmath}
\usepackage{amssymb}
\usepackage{bm}
\usepackage{textcomp}
\usepackage{enumitem}
\usepackage[misc]{ifsym} %\Letter

\usepackage[pagebackref=true,breaklinks=true,letterpaper=true,colorlinks,bookmarks=false]{hyperref}
\usepackage[numbers]{natbib}

\usepackage{caption}
\usepackage{booktabs}
\usepackage{comment}
\usepackage{multirow}
\usepackage{color,xcolor}
\usepackage{wrapfig}
\usepackage{footmisc}

\definecolor{mygreen}{rgb}{0.09, 0.45, 0.27}

\usepackage{multirow}
\usepackage[pagebackref=true,breaklinks=true,colorlinks,bookmarks=false]{hyperref}
\usepackage{arydshln} % \hdashline
\usepackage{orcidlink}
\usepackage{pdfrender}

\usepackage{wrapfig}
\usepackage{subfig}

\usepackage{bm}
\usepackage{colortbl} % \cellcolor
\definecolor{mygray}{gray}{0.6}
\definecolor{mygray-bg}{gray}{0.9}
\definecolor{mygray-u}{gray}{0.92}

\newcommand{\ie}{\textit{i}.\textit{e}.}
\newcommand{\eg}{\textit{e}.\textit{g}.}
\newcommand{\cf}{\textit{cf.}}
\newcommand{\etal}{\textit{et}.\textit{al}.}

\begin{document}
\title{Learning Combinatorial Prompts for Universal Controllable Image Captioning%\thanks{Grants or other notes
}
% \subtitle{Do you have a subtitle?\\ If so, write it here}


\author{Zhen Wang, Jun Xiao, Yueting Zhuang, Fei Gao, Jian Shao and Long Chen$^\textrm{\Letter}$}


\institute{Zhen Wang \at
              College of Computer Science \\ 
              Zhejiang University, Hangzhou, China \\
              \email{zju\_wangzhen@zju.edu.cn}           %  \\
          \and
          Jun Xiao \at
              College of Computer Science \\
              Zhejiang University, Hangzhou, China \\
              \email{junx@cs.zju.edu.cn}
          \and
          Yueting Zhuang \at
              College of Computer Science \\
              Zhejiang University, Hangzhou, China \\
              \email{yzhuang@zju.edu.cn}
          \and
            Fei Gao \at
              College of Computer Science and Technology \\
              Zhejiang University of Technology, Hangzhou, China \\
              \email{feig@zjut.edu.cn}
          \and
         Jian Shao \at
              College of Computer Science \\
              Zhejiang University, Hangzhou, China \\
              \email{jshao@zju.edu.cn}
          \and
          Long Chen \at
              Department of Computer Science and Engineering \\
              The Hong Kong University of Science and Technology, Hong Kong \\
              \email{longchen@ust.hk} \\
        $^\textrm{\Letter}$Long Chen is the corresponding author.
}

\date{Received: date / Accepted: date}

\maketitle
\begin{abstract}
Controllable Image Captioning (CIC) --- generating natural language descriptions about images under the guidance of given control signals --- is one of the most promising directions toward next-generation captioning systems. Till now, various kinds of control signals for CIC have been proposed, ranging from content-related control to structure-related control. However, due to the format and target gaps of different control signals, all existing CIC works (or architectures) only focus on one certain control signal, and overlook the human-like combinatorial ability. By ``combinatorial", we mean that our humans can easily meet multiple needs (or constraints) simultaneously when generating descriptions. To this end, we propose a novel prompt-based framework for CIC by learning \textbf{Com}binatorial \textbf{Pro}mpts, du\-bbed as \textbf{ComPro}. Specifically, we directly utilize a pretrain\-ed language model GPT-2~\cite{radford2019language} as our language model, which can help to bridge the gap between different signal-specific CIC architectures. Then, we reformulate the CIC as a prompt-guide sentence generation problem, and propose a new light\-weight prompt generation network to generate the combinatorial prompts for different kinds of control signals. For different control signals, we further design a new mask attention mechanism to realize the prompt-based CIC. Due to its simplicity, our ComPro can be further extended to more kinds of combined control signals by concatenating these prompts. Extensive experiments on two prevalent CIC benchmarks have verified the effectiveness and efficiency of our ComPro on both single and combined control signals.
\end{abstract}

\section{Introduction}
\label{sec:intro}
\begin{figure*}[t]
    \centering
    \includegraphics[width=\linewidth]{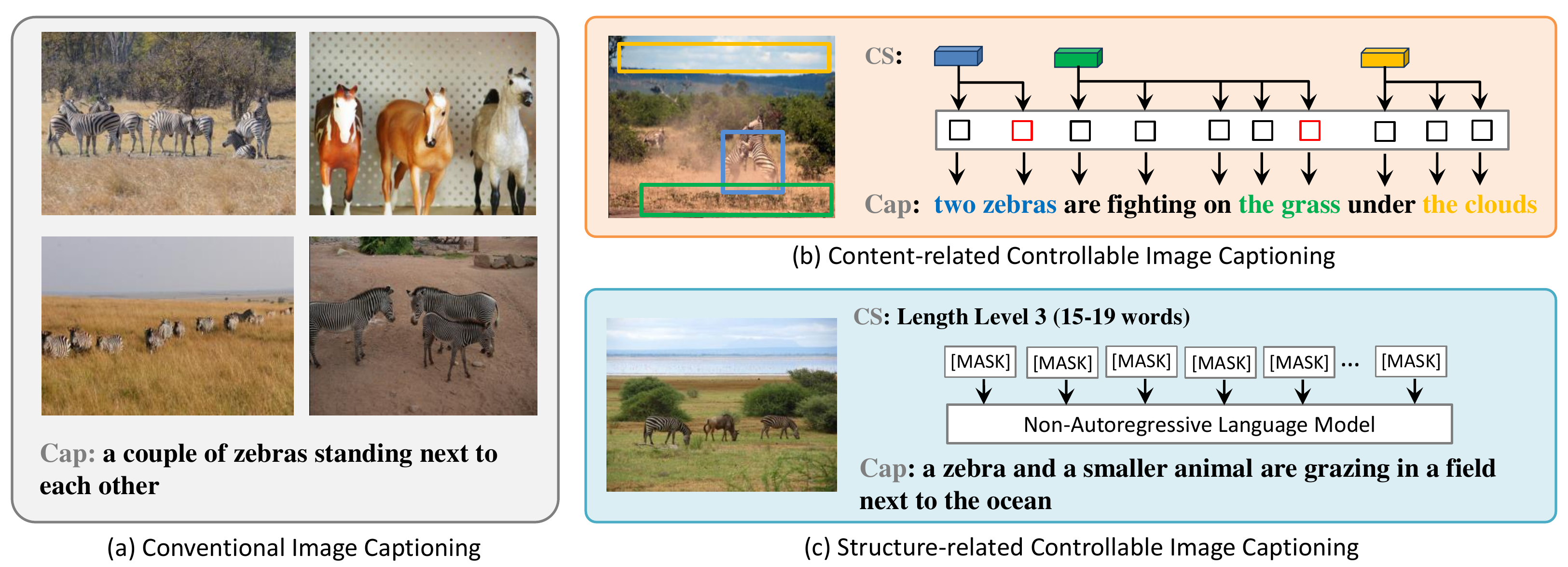}
    % \vspace{-2em}
     \caption{The \textbf{CS} and \textbf{Cap} denote the control signals and generated captions, respectively. \textbf{(a) Conventional Image Captioning}: Existing SOTA captioning models~{\cite{anderson2018bottom}} always generate (almost) the same captions for those images containing similar scenes. \textbf{(b) Content-related CIC}: When the control signal is a sequence of visual regions, the content-related CIC model SCT~\cite{cornia2019show} describes the regions by orders with an extra module to decide the shifting between each region (the \textcolor{red}{red} tokens denote the shifting steps). \textbf{(c) Structure-related CIC}: When the control signal is the length-level, the structure-related CIC model LaBERT~\cite{deng2020length} generates captions with a non-autoregressive model.}
    \label{fig:motivation}
\end{figure*}

Image captioning aims to generate natural language sentences to describe the content of given images~\cite{anderson2018bottom,vinyals2015show,xu2015show,chen2017sca}. However, today's captioning models still tend to generate generic descriptions~\cite{dai2017towards,wang2020compare}. For example, as shown in Fig.~\ref{fig:motivation}(a), for these images that contain similar scenes (\eg, a couple of zebras), their generated captions are exactly the same (\ie, \texttt{a couple of zebras standing next to each other}). Therefore, a surge of recent studies in caption generation focus on generating sentences under the guidance of control signals, called Controllable Image Captioning (\textbf{CIC})~\cite{zheng2019intention,cornia2019show,lindh2020language,chen2020say,zhong2020comprehensive,chen2021human,deshpande2019fast,deng2020length}. In summary, CIC models aim to not only accurately describe the images, but also make their descriptions meet these given constraints.

Currently, existing prevalent control signals for CIC can be mainly categorized into two groups: 1) \textbf{Content-related control signals}: They aim to guide the descriptions with image contents of interest, which are generally formed as a sequence or a set of contents, \eg, objects~\cite{zheng2019intention}, image regions~\cite{cornia2019show,lindh2020language}, scene graphs~\cite{chen2020say,zhong2020comprehensive}, and semantic roles~\cite{chen2021human}. As shown in Fig.~\ref{fig:motivation}(b), these content-related CIC models always generate words based on the sequential control signals. Meanwhile, they need a shifting mechanism to guide the shifting between different content signals. 2) \textbf{Structure-related control signals}: They aim to guide the description with specific sentence structures, which are usually divided into a fixed number of structure types, \eg, different part-of-speech (POS) tag sequences~\cite{deshpande2019fast} or length levels~\cite{deng2020length}. Due to the gaps of different control signals, structure-related CIC models always have different generation pipelines for each specific control signal, \eg, length-controlled CIC models are always non-autoregressive models (\cf~Fig.~\ref{fig:motivation}(c)). 

Thus, due to the signal-specific design purpose in existing CIC works, for different control signals, there are apparent gaps in their captioning model architectures. However, it is worth noting that our humans can always easily combine multiple needs when generating personalized descriptions for an image. For example, when facing some combinatorial constraints, such as ``\texttt{\emph{describing the three objects by orders within nine words}}", our h\-umans are always able to generate qualified sentences for this \emph{combined controls} on both content and sentence structures. In this paper, we argue that all existing CIC frameworks have overlooked such \textbf{\emph{combinatorial ability}}\footnote{By ``combinatorial", we mean that models have strong generalization abilities in two cases: 1) \textbf{Multiple Control Types}: Handling multiple control types simultaneously. 2) \textbf{New Control Instances}: Handling new combinations of control instances which are unseen in the training.} due to the aforementioned signal-specific design nature, and obviously, this ability is essential for real-world applications.

To this end, we propose a novel prompt-based framework for universal controllable image captioning by learning \textbf{Com}binatorial \textbf{Pro}mpts, dubbed as \textbf{ComPro}. By ``universal", we mean that ComPro can efficiently achieve both single and combined controls under a \emph{universal} framework. Specifically, to realize this universal design target, we utilize a pretrained language model GPT-2~{\cite{radford2019language}} as our language model. Due to the extraordinary expression capacity of the GPT model, we can avoid designing signal-specific generation processes or extra modules. Then, we reformulate the CIC task as a prompt-guide sentence generation problem, and propose a new lightweight prompt generation network (PGN) to generate the combinatorial prompts for different kinds of control signals, including either content-related controls, structure-related controls, or combined controls. As the first prompt-based CIC work, we further carefully design specific mask attention mechanisms for both two main types of control signals (\ie, content-related and structure-related), which can be easily extended to combined controls. Meanwhile, by such a universal sub-prompt concatenation design, our ComPro can be further extended to more control signals by generating more sub-prompts. Last but not least, by building on top of the pretrained language model, our ComPro framework can significantly save the number of trainable parameters (\ie, GPT-2 keeps fixed during both training and inference stages).

% We can then efficiently generate the compositional prompt by concatenating these two sub-prompts together.

% \textcolor{red}{includes the classic structure-related control signal sentence length level and the content-related control signal region sequence.} As for sentence length level, we generate different control prefixes based on different length levels and one global region prefix for the entire image. While for region sequence, we generate one global control prefix and different region prefixes for different regions. The control prefix will be concatenated by the region prefix as the final input prefix for GPT-2. Thus, we can efficiently combine these two control signals by concatenating the length-level-specific control prefix and the region-sequence-specific region prefix.

We evaluate our ComPro on two challenging CIC benchmarks: COCO Entities~{\cite{cornia2019show}} and Flickr30K Entities~\cite{plummer2015flickr30k}. Extensive experiments have demonstrated that our ComPro can achieve comparable performance with state-of-the-art signal-specific CIC models on each single control signal, and realize combinatorial ability under combined controls.

In summary, we make three contributions in this paper:

% \vspace{-0.6em}
\begin{enumerate}[leftmargin=4mm]
     \item To the best of our knowledge, ComPro is the first prompt-based CIC model, which can benefit from the pretrained language model with a much efficient training process.
     
     \item ComPro is also the first universal CIC model which can achieve state-of-the-art performance under both content-related and structure-related control signals.

     \item ComPro is the first CIC model with strong combinatorial ability, which can be further extended to new control signals.
\end{enumerate}

%-------------------------------------------------------------------------

\section{Related Work} \label{sec:related_work}

\noindent\textbf{Controllable Image Captioning (CIC).}
Different from traditional image captioning models~\cite{anderson2018bottom,vinyals2015show,xu2015show,chen2017sca} that lack controllability, CIC models aim to accurately describe the given images by taking extra control signals as constraints. Apart from early style-related control signals~\cite{mathews2016senticap,mathews2018semstyle,chen2018factual,gan2017stylenet,shuster2019engaging} that focus on the linguistic style of the generated captions, current mainstream control signals are mainly: 1) \emph{content-related control signals}~\cite{zheng2019intention,cornia2019show,lindh2020language,chen2020say,zhong2020comprehensive,chen2021human} which focus on contents of interest, or 2) \emph{structure-related control signals}~\cite{deshpande2019fast,deng2020length} which focus on the semantic structures. Meanwhile, due to different purposes, these control signals are generally designed with different frameworks and generation processes. Thus, the human-like combinatorial ability (with a universal model) remains challenging. In this paper, we propose the first framework which can achieve good performance under both single and combined control signals.

\noindent\textbf{Prompt Learning with Pretrained Models.} 
Prompt learning or tuning aims to adapt large-scale pretrained language models~\cite{radford2018improving,radford2019language,brown2020language} or visual-language models~\cite{radford2021learning,li2022blip} 
to different downstream tasks by utilizing task-specific prompts as the input. By leveraging these powerful off-the-shelf pretrained big models, current prompt learning works mainly focus on the lightweight task-specific optimization~{\cite{liu2021pre,sun2021ernie,schick2020exploiting,su2022language,chen2022visualgpt}}. Compared to fine-tuning or re-training methods, prompt learning is much more efficient, and can greatly save the training cost. In this paper, we introduce prompt learning into CIC, which can efficiently achieve human-like controllability with significantly fewer trainable parameters.

\noindent\textbf{Prompt Learning for Image Captioning.} Due to the efficient nature of prompt learning, some recent image captioning models also resort to prompt learning~\cite{chen2022visualgpt,luo2022vc,luo2022tuning,mokady2021clipcap}. Specifically, they always steer the pretrained language models with the visual information of images to generate captions. For example, both VC-GPT~\cite{luo2022vc} and I-Tuning~\cite{luo2022tuning} glue the single-modal pretrained image encoder and GPT-2 for image captioning, by only introducing some extra fusion or tuning modules. Also, ClipCap~\cite{mokady2021clipcap} first transforms each image into a fix-length prefix embedding by a mapping network, and then steers or tunes the GPT-2. It is worth noting that all these works either require extra vision-and-language pre-training or need to tune each layer of the GPT-2, \ie, they still suffer high computational costs. In contrast, our ComPro can achieve remarkable performance while the language model is fixed during the whole training time, and we only train the lightweight prompt generation network.

\noindent\textbf{Prompt Learning for Controllable Text Generation.}
As one of the most related tasks of CIC, controllable text generation~\cite{kikuchi2016controlling,keskar2019ctrl} aims to incorporate specific guidance beyond the input text into the language model. Prefix-Tuning~\cite{li2021prefix} is the first work, which proposed to control the sentence generation process of the language models by optimizing a sequence of continuous vectors (named ``prefix"). And these prefixes act as the prepended hidden states of the language models to steer sentence generation. Earlier works mainly focus on learning task-specific prefixes for different tasks, \eg, table-to-text, summarization, or translation~\cite{li2021prefix,lester2021power,tang2022context}. Later, these prompt-based models try to generate attribute-specific prefixes, such as different sentiments or topics.

To further equip the models with similar compositional ability, recent works also combine different prompts to accomplish multi-task or multi-attribute controls~\cite{liu2021pre,han2021ptr,yang2022tailor}. However, all existing works only compose \emph{global coarse-grained} attributes with quite limited choices (\eg, sentiments or topics). In contrast, ComPro can efficiently compose multiple \emph{fine-grained} control signals.

\begin{figure*}[t]
    \centering
    \includegraphics[width=\linewidth]{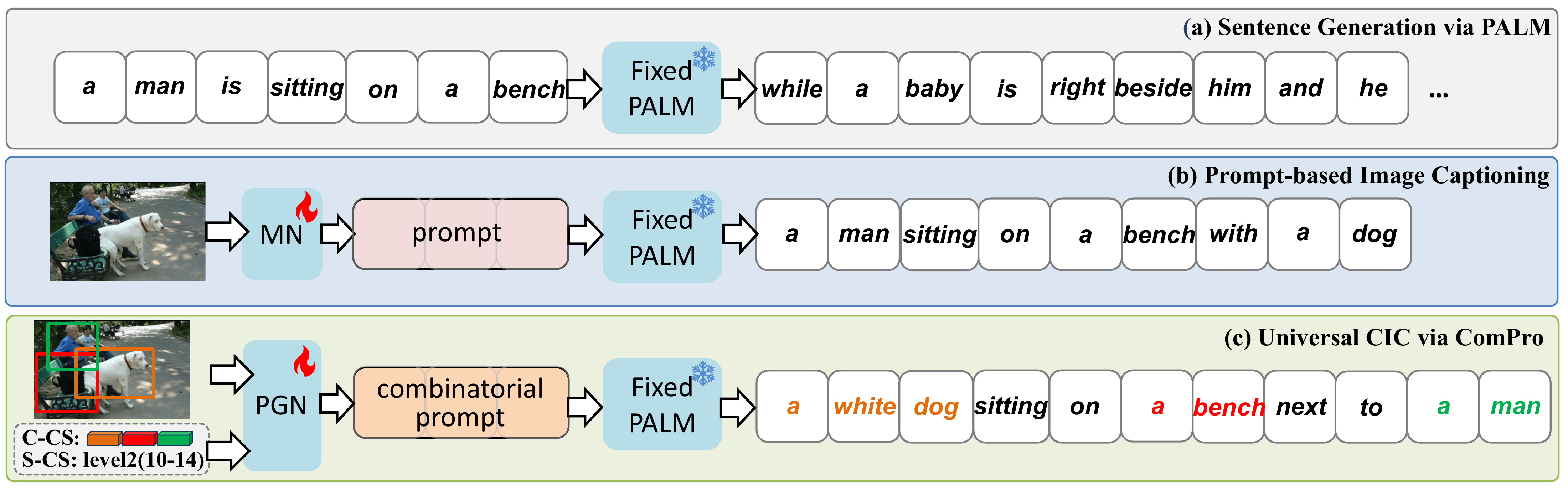}
    % \vspace{-2em}
    \caption{\textbf{(a)} Overview of the sentence generation process via a Pretrained autoregressive Language Model (PLM). \textbf{(b)} Overview of the existing prompt-based image captioning~\cite{mokady2021clipcap}. \textbf{(c)} Overview of our universal controllable image captioning via ComPro.}
    % \vspace{-1em}
    \label{fig:frameworks}
\end{figure*}

% potential of the composition of prompts, there have been several works~\cite{liu2021pre,han2021ptr,yang2022tailor} proposed to combine different sub-prompts to accomplish multi-task or multi-attribute controls. Typically, 
% % the PTR~\cite{han2021ptr} proposed to compose two kinds of manual sub-prompts by logic rules as a multi-task prompt. 
% the Tailor~\cite{yang2022tailor} aims to generate sentences containing multi attributes by simply concatenating single-attribute prompts as a multi-attribute prompt, and got considerable performance. However, existing composition works mainly focus on composing coarse-grained attributes (\eg, sentiment, topic) to control the generation globally, and the class of attributes are rarely small. In this paper, our proposed framework can efficiently compose fine-grained content control signals and structure control signals for each image individually, which includes various composition types and orders.

\section{Method}\label{sec:method}

To equip CIC models with combinatorial ability, we propose a new \textbf{ComPro} framework, which can generate high-quality captions under either single or combined control signals with a universal process. 

In this section, we first introduce the preliminaries about the prompt-based image captioning in Sec.~\ref{sec:prompt_ic}. Then, we introduce the detailed generation process of combinatorial prompts for different control signals, and each component of ComPro in Sec.~\ref{sec:compo_gen} and Sec.~\ref{sec:compo_uni}, respectively. At last, we demonstrate the details of all training objectives and the inference process in Sec.~\ref{sec:train_infer}.

\subsection{Preliminaries: Prompt-based Captioning}
\label{sec:prompt_ic}

\noindent\textbf{Sentence Generation via PLM.} Benefiting from the large amounts of text data, Pretrained autoregressive Language Models (\textbf{PLMs}) have demonstrated a remarkable language generation ability to generate rich and diverse texts. As shown in Fig.~\ref{fig:frameworks}(a), the PLM predicts one token at each time step conditioned on the previous tokens to generate a whole sentence $ \bm{y} = \{y_1, \dots,y_T\}$, \ie, they model the following word probability sequentially: $p(y_t | y_1, \dots,y_{t-1})$.

\noindent\textbf{Prompt-based Image Captioning.} Inspired by the advantages of sentence generation via PLM, some recent image captioning works~\cite{mokady2021clipcap} also try to generate captions in a similar manner. Specifically, as shown in Fig.~\ref{fig:frameworks}(b), they first use a mapping network (MN) to map the input image into a prompt $\bm{P}$, which consists of a set of continuous vectors, and each vector has the same dimension as the word embeddings for PLM. This prompt acts as a sequence of prepended word embeddings contained with visual information of the image to steer the generation of PLM. Then, the PLM can predict the caption tokens starting from the prompt, and model the following probability sequentially to generate the whole caption: $p(y_t | \bm{P},y_1, \dots,y_{t-1})$.

\subsection{Combinatorial Prompt Generation} \label{sec:compo_gen}

Obviously, the key to the success of prompt-based image captioning lies in the way of generating specific prompts for each image, \ie, translating the visual information to the same textual semantic space of PLM. As shown in Fig.~\ref{fig:frameworks}(c), to achieve universal controllable image captioning, we propose a prompt generation network (PGN) to learn \emph{combinatorial prompts}, which consists of the information from both input image and all different control signals. In this manner, we hope this combinatorial prompt can guide the generation of PLM to describe the image content while the sentences meet the content or structure constraints.

\begin{figure}[t]
    \centering
    \includegraphics[width=0.9\linewidth]{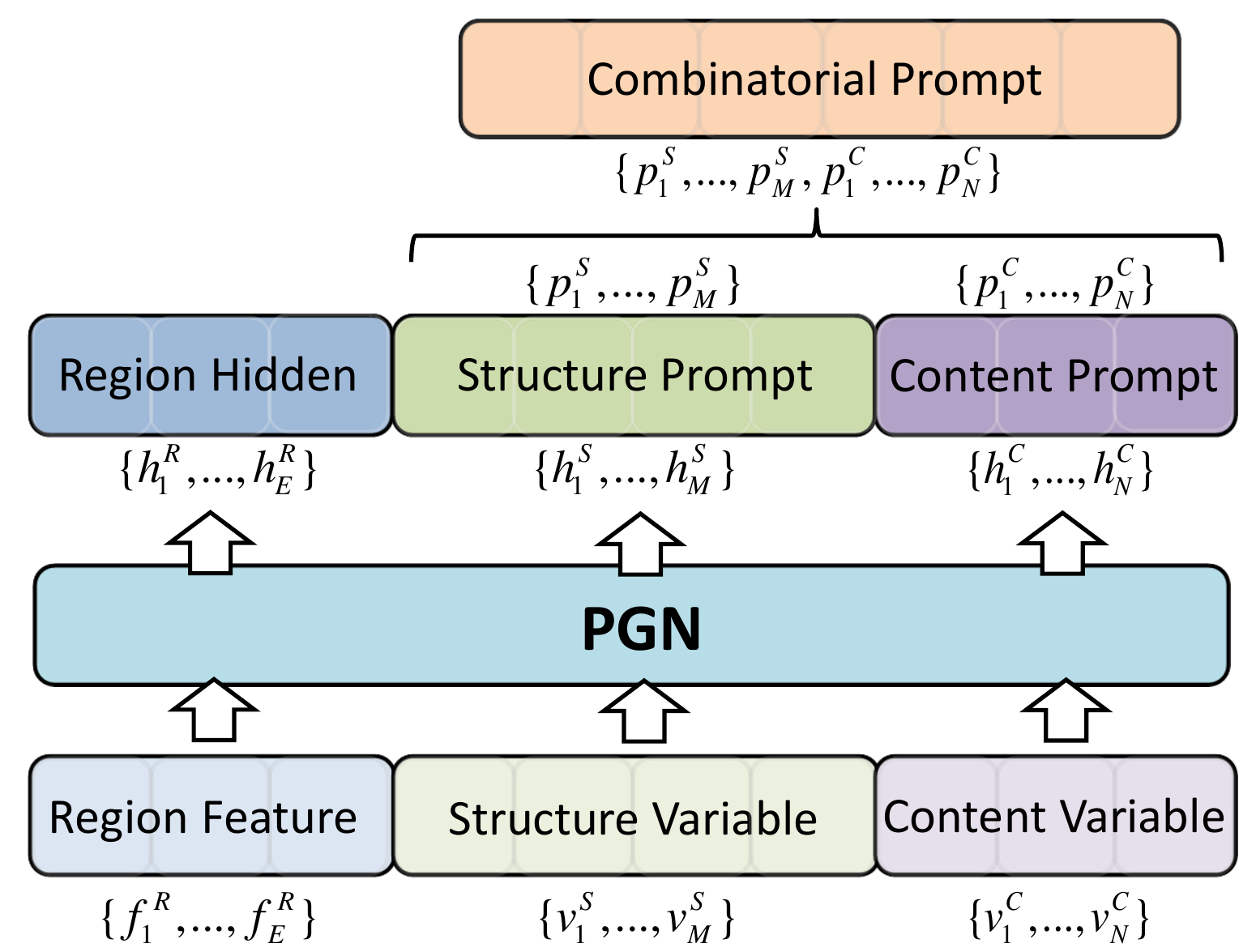}
    % \vspace{-1.0em}
    \caption{Overview of the combinatorial prompt generation process.}
   % \vspace{-1.0em}
    \label{fig:pgn}
\end{figure}

\noindent\textbf{Format of Combinatorial Prompt.} Since we mainly discuss the combined control signals from two types of constraints (\ie, content-/structure-related), we first show the combinatorial prompts here are composed of two kinds of sub-prompts.
% \footnote{Although we only discuss two sub-prompts here, our framework can also be easily extended to more complex situations with more constraints.}. 
Specifically, we call the two sub-prompts as \textbf{\emph{content prompt}} $\bm{p}^{C}=\{p^{C}_1, \dots,p^{C}_N\} \in\mathbb{R}^{N*D}$ and \textbf{\emph{structure prompt}} $\bm{p}^{S}=\{p^{S}_1, \dots,p^{S}_M\} \in\mathbb{R}^{M*D}$, where $M$ and $N$ are prompt lengths, and $D$ is the hidden dimension same to word embedding for PLM. Thus, combinatorial prompt $\bm{P} \in \mathbb{R}^{(MK+NJ)*D}$ is denoted as: % the concatenation of them:
\begin{equation} \label{eq:1}
    % \bm{p} = [\bm{P}^{S}; \bm{P}^{C}] = \{p^{S}_1, \dots,p^{S}_M,p^{C}_1, \dots,p^{C}_N\}. 
    \bm{P} = [\bm{P}^{S}; \bm{P}^{C}]= [\bm{p}^{S}_1; \dots;\bm{p}^{S}_K;\bm{p}^{C}_1; \dots;\bm{p}^{C}_J]. 
\end{equation}
where $K$ and $J$ are prompt numbers, as each control type may consist of multiple prompts (\eg, $J$ prompts for the same content control, $\bm{P}^{C} = \{\bm{p}^{C}_1,\dots, \bm{p}^{C}_J\}$).

\begin{figure*}[t]
  \centering
  \includegraphics[width=1.0\linewidth]{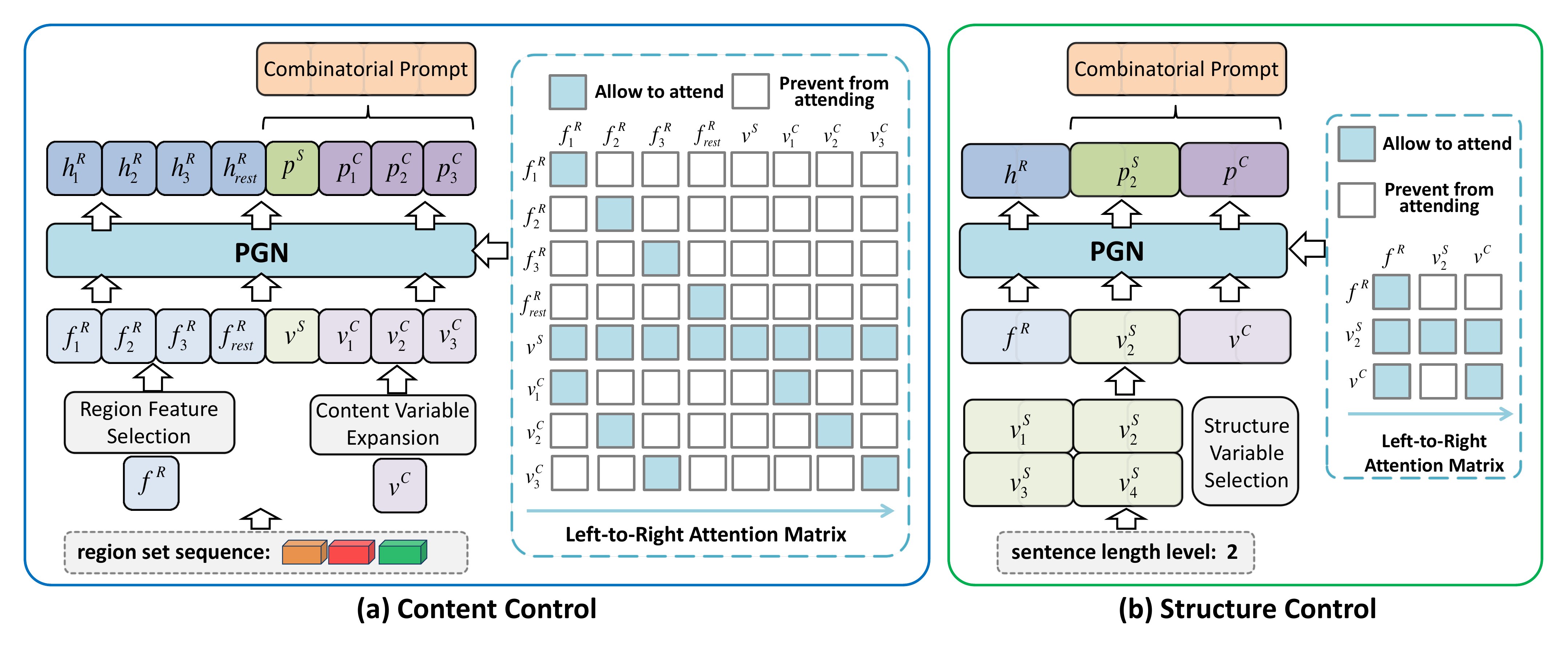}
  % \vspace{-2em}
  \caption{The pipeline of different combinatorial prompt generation: \textbf{(a) Content-related control}: It takes a region set sequence with three object sets as control signals. \textbf{(b) Structure-related control}: It takes the length level-2 as the control signal (four levels in total).}
  % \vspace{-1.0em}
  \label{fig:different_controls}
\end{figure*}

\noindent\textbf{\emph{Extension.}} {For simplicity, we only introduce the simplest case in the method section (\ie, two control types). Beyond this, we can also be easily extended to more new control types by generating and concatenating more sub-prompts:}
\begin{equation}
    \bm{P} = [\bm{P}^{S_1}; \dots; \bm{P}^{S_A}; \bm{P}^{C_1}; \dots; \bm{P}^{C_B}],
\end{equation}
where $[;]$ is the concatenation operation, $A$ and $B$ are the numbers of structure- and content-related control types.

\noindent\textbf{Prompt Generation Network (PGN).} The prompt generation network (PGN) is constructed based on the standard Transformer~\cite{vaswani2017attention} architecture, which has strong context and representation encoding abilities. Specifically, we utilize the standard Transformer encoder blocks with multi-head self-attention. Meanwhile, to learn meaningful and targeted information for the combinatorial prompt, we further propose a new \emph{signal-specific mask attention mechanism} for the multi-head attention conditioned on different control signals (More details are shown in Sec.~\ref{sec:compo_uni} and Fig.~\ref{fig:different_controls}).

As shown in Fig.~\ref{fig:pgn}, given an input image $I$, PGN takes the region features $\bm{R} = \{r_1, \dots,r_E\}$ of the image and two sets of learnable variables, the content variables $\bm{v}^{C} = \{v^{C}_1, \dots,v^{C}_N\}$ and structure variables $\bm{v}^{S} = \{v^{S}_1, \dots,v^{S}_M\}$, as inputs. Specifically, $R$ is a set of visual features of the corresponding regions in $I$ extracted by an object detector. Both $\bm{v}^{C}\in \mathbb{R}^{N*D}$ and $\bm{v}^{S} \in \mathbb{R}^{M*D}$ have the same hidden dimension and length with the content and structure prompt, respectively. We generate the combinatorial prompt $\bm{P}$ as:
\begin{equation}
\begin{gathered}
\text{input} = [\mathtt{MLP}(\bm{R});\bm{v}^{S};\bm{v}^{C}], \\
[\bm{h}^{R}; \bm{h}^{S}; \bm{h}^{C}] = \mathtt{PGN}(\text{input}), \\
\bm{p}^{S} = \bm{h}^{S}, \bm{p}^{C} = \bm{h}^{C}, \bm{P} = [\bm{p}^{S};\bm{p}^{C}],
\end{gathered}    
\end{equation}
where $\mathtt{MLP}$ is a learnable MLP to project region features $\bm{r}$ to the PLM space as $\bm{f}^{R} = \{f^{R}_1, \dots,f^{R}_E\} \in \mathbb{R}^{E*D}$. The hidden states of content variables $\bm{h}^{C}=\{h^{C}_1, \dots,h^{C}_N\}$ and structure variables $\bm{h}^{S}=\{h^{S}_1, \dots,h^{S}_M\}$ are regarded as the content prompt $\bm{p}^C$ and structure prompt $\bm{p}^S$, respectively. Finally, we obtain combinatorial prompt $\bm{P}$ by concatenating these two sub-prompts (we only show $K=J=1$ here).

\subsection{Combinatorial Prompt for Universal Controls} 
\label{sec:compo_uni}

To achieve universal CIC, we introduce the detailed generation process and proposed signal-specific mask attention mechanisms for different control signals, including content control, structure control, and combine control.

\subsubsection{Combinatorial Prompt for Content Control}
Given the content control signal as an ordered sequence of image contents, \eg, the classic region set sequence $\bm{R}^{\text{seq}} = \{\bm{r}^{\text{set}}_{1},\dots,\bm{r}^{\text{set}}_{Q}\}$ ($Q$ is region set number), our goal is to generate the caption which in turn describes all the regions. Particularly, each region set contains one or more regions with the same object class. Thus, we aim to generate an ordered sequence of content prompts and let each content prompt \emph{``know"} its corresponding control contents.

\textbf{Region Feature Selection.}
As shown in Fig.~\ref{fig:different_controls}(a), we first select the corresponding region features $\bm{f}^{R}_{i}$ for each region set $\bm{r}^{\text{set}}_{i}$, then we can re-organize all region features as: $\bm{F}^{R} = \{\bm{f}^{R}_{1},\dots,\bm{f}^{R}_{Q},\bm{f}^{R}_{\text{rest}}\}$, where $\bm{f}^{R}_{\text{rest}}$ are the region features for all the rest regions.

\textbf{Content Variable Expansion.} Meanwhile, we copy the content variables $\bm{v}^{C}$ for $Q$ times as a content variable sequence $ \bm{V}^{C}  = \{\bm{v}^{C}_1,\dots,\bm{v}^{C}_Q\}$, where each content variable $\bm{v}^{C}_i$ corresponds to one region set. We then concatenate all region features and variables together to obtain the input:
\begin{equation}
\text{input}_c =[\bm{f}^{R}_{1};\dots;\bm{f}^{R}_{Q};\bm{f}^{R}_{\text{rest}};\bm{v}^{S};\bm{v}^{C}_1;\dots;\bm{v}^{C}_Q],
\end{equation}
where $\bm{v}^S$ are the learnable structure variables.

\textbf{Content-Control Mask Attention.} To realize the content control target, we design a specific mask attention mechanism. As the attention matrix shown in Fig.~\ref{fig:different_controls}(a), the design philosophy is as follows: 1) The region features of each region set can only attend to themselves to learn better object information within each set. 2) Each content variable in each $\bm{v}^{C}_i$ can attend to themselves and their corresponding region set features $\bm{f}^{R}_i$ to learn visual information and control guidance of each region set. 3) The structure variables $\bm{v}^{S}$ can attend to themselves, all region features, and content variables to learn the global control information and visual information among all regions. 

Then, we can generate the combinatorial prompt $\bm{P}$ by:
\begin{equation}
\begin{gathered}
\relax [\bm{h}^{R}_{1};\dots;\bm{h}^{R}_{Q};\bm{h}^{R}_{\text{rest}};\bm{h}^{S};\bm{h}^{C}_1;\dots;\bm{h}^{C}_Q] = \mathtt{PGN}(\text{input}_c), \\
\bm{p}^{S} = \bm{h}^{S}, \bm{p}^{C}_1=\bm{h}^{C}_1,\dots,\bm{p}^{C}_Q = \bm{h}^{C}_Q, \\
\bm{P} = [\bm{p}^{S}; \bm{p}^{C}_1; \dots; \bm{p}^{C}_Q].
\end{gathered}
\end{equation}

\begin{figure}[t]
    \centering
    \includegraphics[width=0.7\linewidth]
    {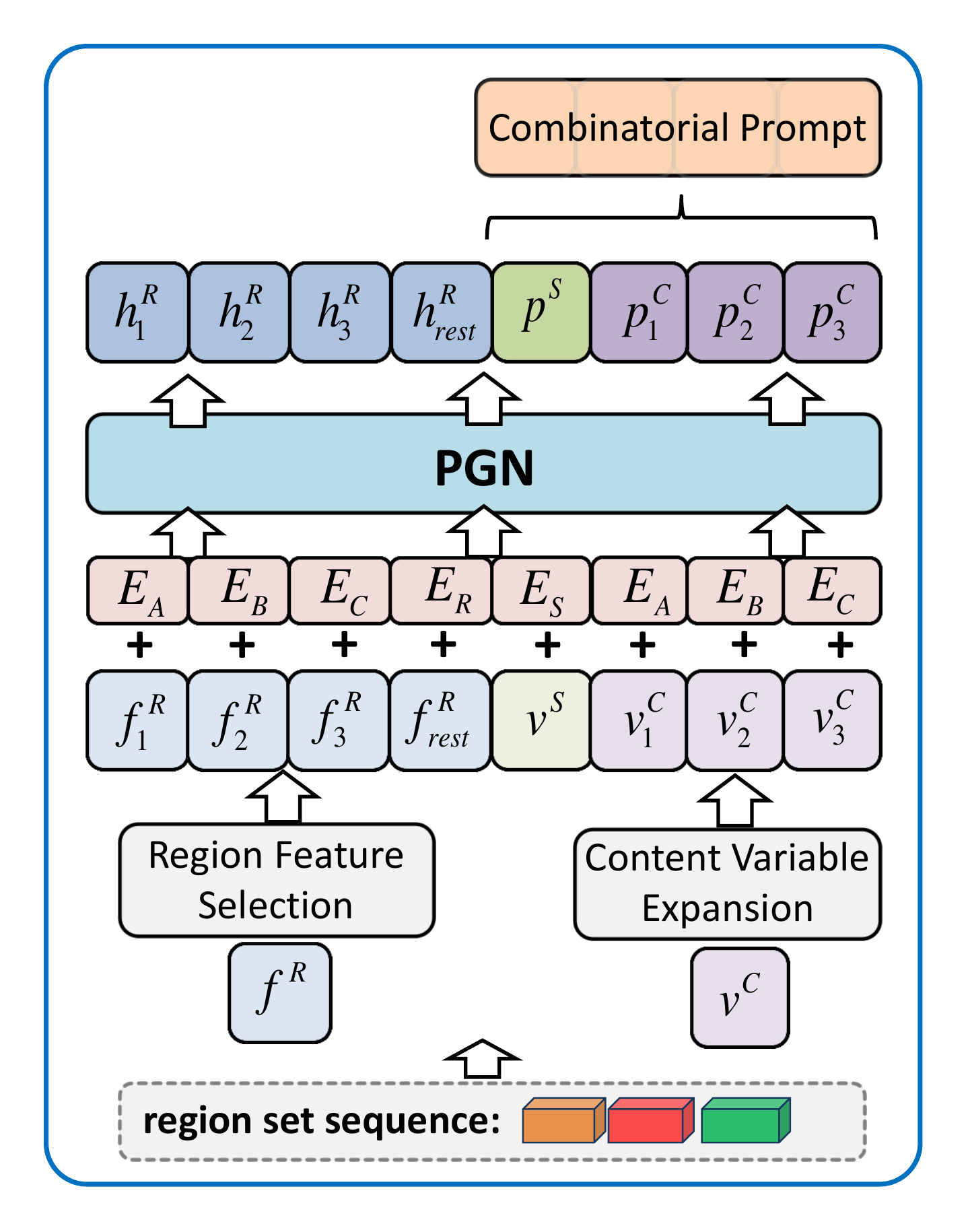}
    % \vspace{-1em}
    \caption{The pipeline of combinatorial prompt generation for content control with specific content-control segment embedding. It takes a region set sequence with three objects as the content control signal.}
    % \vspace{-1.5em}
    \label{fig:pgn_seg}
\end{figure}

\noindent\textbf{Discussion.} Apart from mask attention, we can also apply a specific content-control segment embedding to the input to help content variables learn control guidance of each region set. As shown in Fig.~\ref{fig:pgn_seg}, each content variable $\bm{v}^{C}_i$ and their corresponding region set features $\bm{f}^{R}_i$ share the same type of segment embedding $E$. Detailed comparisons and ablations are in Sec.~\ref{sec:ablation}.

\subsubsection{Combinatorial Prompt for Structure Control}

Given the structure control signal such as different sentence length levels, our goal is to generate captions that meet these sentence structure constraints. Thus, we aim to generate multiple structure prompts and let each structure prompt \emph{``know"} its corresponding control types.

\noindent\textbf{Structure Variable Selection.} As shown in Fig.~\ref{fig:different_controls}(b), we first expand the structure variables $\bm{v}^{S}$ into a set of structure variables $ \bm{V}^{S}  = \{\bm{v}^{S}_1,\dots, \bm{v}^{S}_L\}$, where $L$ is the number of different structure types, such as the number of different length levels. And each structure variable $\bm{v}^{S}_i$ corresponds to one level respectively. Given the image and specific structure type, we first select the corresponding structure variable to obtain the input:
\begin{equation}
\text{input}_s = [\bm{f}^{R}; \bm{v}^{S}_i; \bm{v}^{C}].
\end{equation}

\noindent\textbf{Structure-Control Mask Attention.} Similarly, for realizing the structure control target, we design a specific mask attention mechanism. As the attention matrix shown in Fig.~\ref{fig:different_controls}(b), the design philosophy is as follows: 1) All region features only attend to themselves to encode high-order visual relations and visual contexts. 2) The content variables $\bm{v}^{C}$ can attend to themselves and all region features to catch the global visual information of the image. 3) The structure variables in $\bm{v}^{S}_i$ can attend to all input tokens to learn the global control information.

Then, we can generate the combinatorial prompt $\bm{P}$ by:
\begin{equation}
\begin{gathered}
\relax [\bm{h}^{R}; \bm{h}^{S}_i; \bm{h}^{C}] = \mathtt{PGN}(\text{input}_s), \\
\bm{p}^{S}_i =\bm{h}^{S}_i, \bm{p}^{C} = \bm{h}^{C}, \bm{P} = [\bm{p}^{S}_i;\bm{p}^{C}].
\end{gathered}
\end{equation}

\subsubsection{Combinatorial Prompt for Combined Control}
\label{sec:compo_combine}
Due to the simplicity and generalization ability of ComPro, it can be easily extended to combined control signals. Correspondingly, our goal is not only to describe images, and to make sentences meet both content and structure constraints. We can achieve the combined control by combining the process of each single control together.

\begin{figure*}[t]
    \centering
    \includegraphics[width=0.7\linewidth]{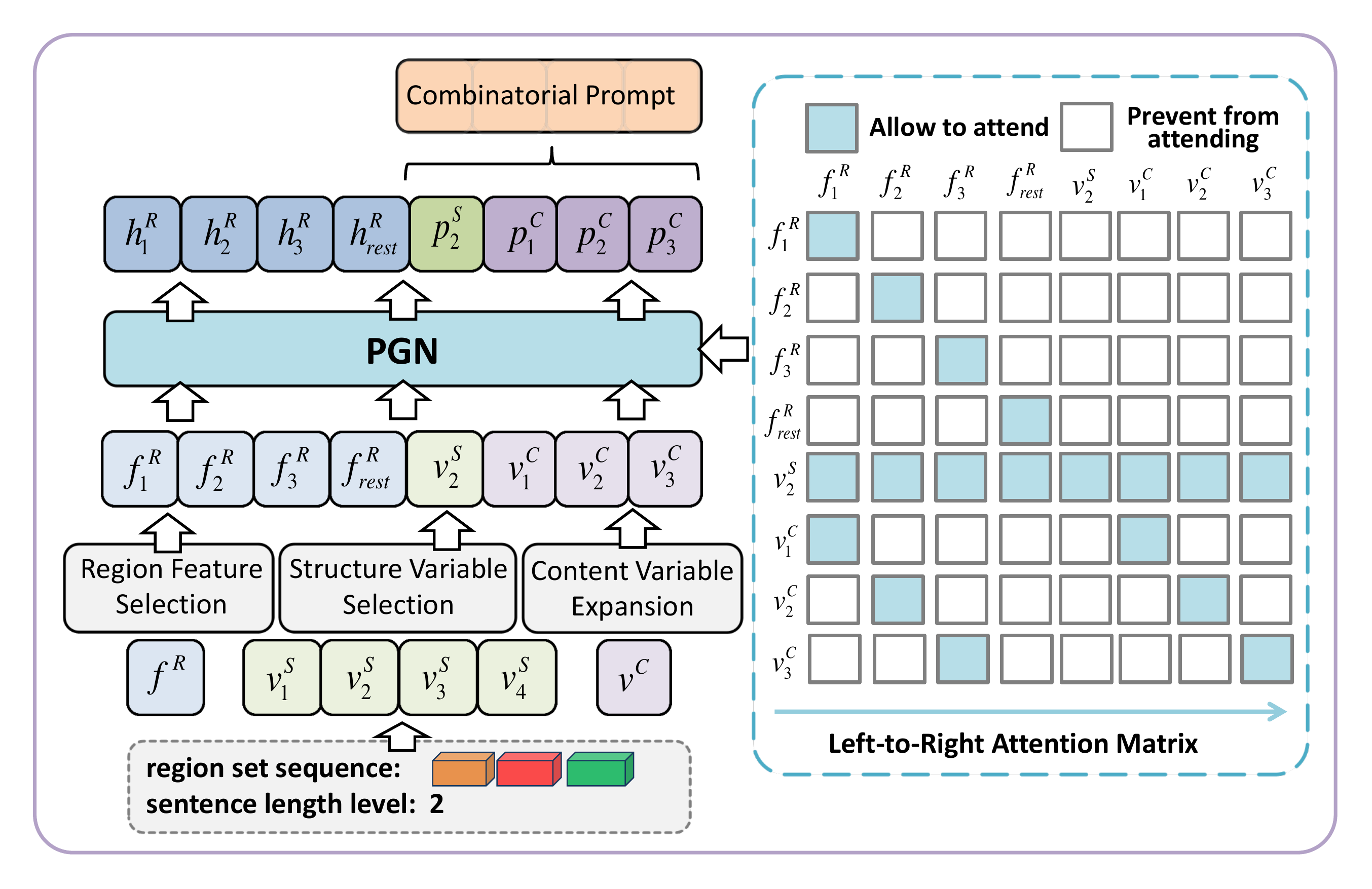}
    % \vspace{-1em}
    \caption{The pipeline of combinatorial prompt generation for combined control on both sentence content and structure. It takes a region set sequence with three objects as the content control signal, and the length level-2 as the structure control signal (four levels in total).}
    % \vspace{-1.5em}
    \label{fig:pgn_combined}
\end{figure*}

\noindent\textbf{Region Feature Selection.}
As shown in Fig~\ref{fig:pgn_combined}, we first select the corresponding region features $\bm{f}^{R}_{i}$ for each region set $\bm{r}^{\text{set}}_{i}$, then we can re-organize all region features as: $\bm{F}^{R} = \{\bm{f}^{R}_{1},\dots,\bm{f}^{R}_{Q},\bm{f}^{R}_{\text{rest}}\}$, where $\bm{f}^{R}_{\text{rest}}$ are the region features for all the rest regions.

\noindent\textbf{Content Variable Expansion.} Meanwhile, we copy the content variables $\bm{v}^{C}$ for $Q$ times as a content variable sequence $ \bm{V}^{C}  = \{\bm{v}^{C}_1,\dots,\bm{v}^{C}_Q\}$, where each content variable $\bm{v}^{C}_i$ corresponds to one region set. 

\noindent\textbf{Structure Variable Selection.} At last, we expand the structure variables $\bm{v}^{S}$ into a set of structure variables $ \bm{V}^{S}  = \{\bm{v}^{S}_1,\dots, \bm{v}^{S}_L\}$, where $L$ is the number of different structure types, such as the number of different length levels. And each structure variable $\bm{v}^{S}_i$ corresponds to one level respectively. Given the specific structure type, we select the corresponding structure variable and concatenate all features and content variables together to obtain the input:
\begin{equation}
\text{input}_{*} =[\bm{f}^{R}_{1};\dots;\bm{f}^{R}_{Q};\bm{f}^{R}_{\text{rest}};\bm{v}^{S}_i;\bm{v}^{C}_1;\dots;\bm{v}^{C}_Q].
\end{equation}

\noindent\textbf{Combined-Control Mask Attention.} As the attention matrix shown in Fig.~\ref{fig:pgn_combined}, the design philosophy is as follows: 1) The region features of each region set can only attend to themselves to learn better object information within each set. 2) Each content variable in each $\bm{v}^{C}_i$ can attend to themselves and their corresponding region set features $\bm{f}^{R}_i$ to learn visual information and control guidance of each region set. 3) The structure variables $\bm{v}^{S}_i$ can attend to themselves, all region features, and content variables to learn the global control information and visual information among all regions. 
We generate the combinatorial prompts $\bm{P}$ for combined control as follows:
\begin{equation}
\begin{gathered}
\relax [\bm{h}^{R}_{1};\dots; \bm{h}^{R}_{Q}; \bm{h}^{R}_{\text{rest}}; \bm{h}^{S}_i; \bm{h}^{C}_1; \dots; \bm{h}^{C}_Q] = \mathtt{PGN}(\text{input}_*), \\
\bm{p}^{S}_i = \bm{h}^{S}_i, \bm{p}^{C}_1 = \bm{h}^{C}_1, \dots, \bm{p}^{C}_Q = \bm{h}^{C}_Q, \\
\bm{P} = [\bm{p}^{S}_i; \bm{p}^{C}_1; \dots; \bm{p}^{C}_Q].
\end{gathered}
\end{equation}

\noindent\textbf{Highlights.} By generating and concatenating more sub-pro\-mpts, ComPro can be further extended to more new control signals beyond two control signals, \eg, for another type of structure signal, we can also train a similar sub-prompt for each different discrete structure type (\cf, Sec.~\ref{sec:4.4.2}).

% \lc{say our method is general enough to extend to other control signals.}
 
\subsection{Training Objectives and Inference}
\label{sec:train_infer}
\noindent\textbf{Training Objectives.} During training, we concatenate the combinatorial prompt to the caption tokens as the prompt-caption concatenation: $\bm{z} = \{\bm{P}, y_1, \dots,y_T\}$.

Our training objective is to predict the caption tokens conditioned on the prompt and previously predicted tokens autoregressively. Since the language model is fixed, we only train the PGN using the cross-entropy loss, \ie,
\begin{equation}
   L_{xe} = - \textstyle{\sum}_{i=1}^T log p_{\theta}(y_t | \bm{P}, y_1, \dots,y_{t-1}).
\end{equation}
where $\theta$ denotes the trainable parameters of PGN.

\noindent\textbf{Inference.}
In inference, firstly, we generate a specific combinatorial prompt based on each image and different control signals by the prompt generation network. Then, we generate the caption conditioned on the combinatorial prompt using the PLM, which predicts the next token one by one to generate the final caption. For each token, the PLM outputs probabilities for all vocabulary tokens, which are used to determine the next one by employing the beam search.

\section{Experiments}
\label{sec:experiments}

\subsection{Experimental Settings}

\textbf{Datasets.} We evaluated our ComPro on two CIC benchmarks: 1) \textbf{\emph{COCO Entities}}~\cite{cornia2019show}: It was built based on the COCO dataset~\cite{lin2014microsoft}, which contains 120,000 images annotated with 5 captions each. Each entity (noun) in the caption is \emph{automatically aligned} to the regions with the same object class extracted by the detector~\cite{ren2015faster}. 2) \textbf{\emph{Flickr30K Entities}}~\cite{plummer2015flickr30k}: It was built based on the Flickr30K dataset~\cite{young2014image}, which contains 31,000 images annotated with 5 captions each. Different from COCO Entities, each entity in the caption is \emph{manually grounded} with one or more regions. In both datasets, we used the same Karpathy splits~\cite{karpathy2015deep}, and followed the setting of prior works~\cite{cornia2019show} to remove the captions with empty region sets from validation and test sets.

\textbf{Control Signal Choices.} Since we are the first CIC work to discuss the combinatorial ability, without loss of generality, we evaluated our methods on two representative control signals: 1) \textbf{\emph{Content control}}: For content-related control signals, we used the region-control proposed by Cornia~\etal~\cite{cornia2019show}. 
This control signal is formed as a region set sequence and each region set contains one or more regions with the same detection object class or manually annotated class. 
CIC models aim to in turn describe all the regions.  2) \textbf{\emph{Structure control}}: For structure-related control signals, we used the length-control proposed by Deng~\etal~\cite{deng2020length}. This control signal is a sentence length level, \ie, CIC models aim to generate captions with lengths inside given ranges. 
% \textcolor{blue}{To further show the generalization ability on more compositional control types, we extended it to more control signals. Specifically, we used the tense-control~\cite{zhu2021self}, which aims to generate captions with given tense types.} 

\textbf{Metrics.}
For the caption quality evaluation, we used two types of evaluation metrics: 1) We followed general image captioning works, and used all the prevalent accuracy-based metrics: BLEU-4 (\textbf{B-4})~\cite{papineni2002bleu}, METEOR (\textbf{M})~\cite{banerjee2005meteor}, ROUGE-L (\textbf{R})~\cite{lin2004rouge}, CIDEr-D (\textbf{C})~\cite{vedantam2015cider}, and SPICE (\textbf{S})~\cite{anderson2016spice}. Particularly, we evaluated generated captions against their single ground-truth caption~\cite{chen2021human}. 2) Meanwhile, following previous CIC works~\cite{cornia2019show,deng2020length}, to further evaluate the controllability of the CIC models, we also employed two different metrics for the two controls. Specifically, for region-control CIC, we used the alignment score based on the Needleman-Wunsch algorithm (\textbf{NW})~\cite{needleman1970general,cornia2019show} to evaluate how the caption follows the region sequence. For length-control CIC, we used the length precision (\textbf{LP})~\cite{deng2020length} to evaluate whether the lengths of the generated captions are within desired length ranges\footnote{We followed existing CIC works and evaluated the controllability together with accuracy-based metrics globally. More reasonable evaluation metrics for different CIC are beyond the scope of this work.}.
% \textcolor{blue}{For tense-control CIC, we used the tense precision (\textbf{TP}) to evaluate whether the tense of the generated captions is of desired tense types.}

\subsection{Implementation Details}

\addtolength{\tabcolsep}{-2pt}
\begin{table*}[t]
    \renewcommand\arraystretch{1.05}
    \centering
    \scalebox{1.06}{
        \begin{tabular}
            {c|l|c|c|ccccc|c|ccccc|c}
            \hline
            \multirow{2}{*}{Datasets} & \multirow{2}{*}{Method} &  \multirow{2}{*}{Size} & \multirow{2}{*}{Params} & \multicolumn{6}{c|}{\textbf{Content-related Control}} & \multicolumn{6}{c}{\textbf{Structure-related Control}} \\
            \cline{5-16}
            & & &  & {B4} & {M} & {R} & {C} & {S} & {NW} & {B4} & {M} & {R} & {C} & {S} & {LP} \\
            \hline
                & SCT~\cite{cornia2019show} & -- & 70M & 19.7 & 24.0 & 51.6 & 192.3 & 45.8 & 0.508 & -- & -- & -- & -- & -- & --  \\
                & LaBERT$^*$~\cite{deng2020length} & -- &  140M & -- & -- & -- & -- &  -- &  -- & 13.5 & 20.6 & \textcolor{blue}{\textbf{42.3}} & 136.6 & \textcolor{blue}{\textbf{32.4}} & \textcolor{red}{99.7}  \\
            % \cline{2-16}
                COCO & \textbf{ComPro (Ours)} & \cellcolor{mygray-bg}{B} & \cellcolor{mygray-bg}{39M} & \textcolor{red}{\cellcolor{mygray-bg}{23.6}} & \cellcolor{mygray-bg}{27.1} & \cellcolor{mygray-bg}{55.6} & \textcolor{red}{\cellcolor{mygray-bg}{227.5}} & \textcolor{red}{\cellcolor{mygray-bg}{49.9}} & \cellcolor{mygray-bg}{0.519} & \cellcolor{mygray-bg}{13.7} & \cellcolor{mygray-bg}{20.3} & \cellcolor{mygray-bg}{41.7} & \cellcolor{mygray-bg}{134.1} & \cellcolor{mygray-bg}{31.7}  & \cellcolor{mygray-bg}{92.4}  \\
                
                Entities & \textbf{ComPro (Ours)}  & \cellcolor{mygray-bg}{M} & \cellcolor{mygray-bg}{69M} & \cellcolor{mygray-bg}{23.5} & \textcolor{red}{\cellcolor{mygray-bg}{27.2}} & \textcolor{red}{\cellcolor{mygray-bg}{55.6}} & \cellcolor{mygray-bg}{226.1} & \cellcolor{mygray-bg}{49.8} & \textcolor{red}{\cellcolor{mygray-bg}{0.520}} & \cellcolor{mygray-bg}{13.6} & \cellcolor{mygray-bg}{20.5} & \cellcolor{mygray-bg}{41.6} & \cellcolor{mygray-bg}{134.8} & \cellcolor{mygray-bg}{31.4} & \cellcolor{mygray-bg}{92.5} \\
                
                & \textbf{ComPro (Ours)} & \cellcolor{mygray-bg}{L} & \cellcolor{mygray-bg}{107M} & \textcolor{blue}{\textbf{\cellcolor{mygray-bg}{24.0}}} & \textcolor{blue}{\textbf{\cellcolor{mygray-bg}{27.3}}} & \textcolor{blue}{\textbf{\cellcolor{mygray-bg}{56.1}}} & \textcolor{blue}{\textbf{\cellcolor{mygray-bg}{232.2}}} & \textcolor{blue}{\textbf{\cellcolor{mygray-bg}{50.4}}} & \textcolor{blue}{\textbf{\cellcolor{mygray-bg}{0.531}}} & \textcolor{red}{\cellcolor{mygray-bg}{13.9}} & \textcolor{blue}{\textbf{\cellcolor{mygray-bg}{20.8}}} & \cellcolor{mygray-bg}{41.9} & \textcolor{red}{\cellcolor{mygray-bg}{138.5}} & \textcolor{red}{\cellcolor{mygray-bg}{32.0}} & \cellcolor{mygray-bg}{94.7} \\

            \cline{2-16}
                & \textbf{ComPro$^{*}$ (Ours)}  & \cellcolor{mygray-bg}{L} & \cellcolor{mygray-bg}{107M}   & \cellcolor{mygray-bg}{--} & \cellcolor{mygray-bg}{--} & \cellcolor{mygray-bg}{--} & \cellcolor{mygray-bg}{--} & \cellcolor{mygray-bg}{--} & \cellcolor{mygray-bg}{--}  & \textcolor{blue}{\textbf{\cellcolor{mygray-bg}{13.9}}} & \textcolor{red}{\cellcolor{mygray-bg}{20.7}} & \textcolor{red}{\cellcolor{mygray-bg}{41.9}} & \textcolor{blue}{\textbf{\cellcolor{mygray-bg}{138.5}}} & \cellcolor{mygray-bg}{31.8} & \textcolor{blue}{\textbf{\cellcolor{mygray-bg}{99.8}}} \\
            \hline
                & SCT~\cite{cornia2019show} & -- & 70M & 9.7 & 14.6 & 35.0 & 72.3 & 23.6 & 0.147 & -- & -- & -- & -- & -- & --  \\
                & LaBERT$^*$~\cite{deng2020length} & -- & 140M & -- & -- & -- & -- &  -- &  -- & 8.1 & 14.6 & 32.7 & 70.8 &  \textcolor{red}{19.6} & \textcolor{blue}{\textbf{98.4}} \\
            % \cline{2-16}
                Flickr30K & \textbf{ComPro (Ours)} & \cellcolor{mygray-bg}{B} & \cellcolor{mygray-bg}{39M} & \cellcolor{mygray-bg}{11.4} & \cellcolor{mygray-bg}{16.5} & \cellcolor{mygray-bg}{37.3} & \cellcolor{mygray-bg}{84.1} & \textcolor{red}{\cellcolor{mygray-bg}{23.7}} & \cellcolor{mygray-bg}{0.195} & \cellcolor{mygray-bg}{9.0} & \cellcolor{mygray-bg}{14.6} & \cellcolor{mygray-bg}{32.7} & \cellcolor{mygray-bg}{67.8} & \cellcolor{mygray-bg}{19.3} & \cellcolor{mygray-bg}{80.1} \\
            
                Entities & \textbf{ComPro (Ours)} & \cellcolor{mygray-bg}{M} & \cellcolor{mygray-bg}{69M} & \textcolor{red}{\cellcolor{mygray-bg}{11.6}} & \textcolor{red}{\cellcolor{mygray-bg}{17.0}} & \textcolor{red}{\cellcolor{mygray-bg}{37.3}} & \textcolor{red}{\cellcolor{mygray-bg}{86.6}} & \cellcolor{mygray-bg}{23.6}  & \textcolor{red}{\cellcolor{mygray-bg}{0.196}} & \cellcolor{mygray-bg}{8.9} & \cellcolor{mygray-bg}{14.7} & \cellcolor{mygray-bg}{32.6} & \cellcolor{mygray-bg}{67.9} & \cellcolor{mygray-bg}{18.9} & \cellcolor{mygray-bg}{81.8}  \\
               
                & \textbf{ComPro (Ours)} & \cellcolor{mygray-bg}{L} & \cellcolor{mygray-bg}{107M} & \textcolor{blue}{\textbf{\cellcolor{mygray-bg}{11.9}}} & \textcolor{blue}{\textbf{\cellcolor{mygray-bg}{17.3}}} & \textcolor{blue}{\textbf{\cellcolor{mygray-bg}{37.8}}} & \textcolor{blue}{\textbf{\cellcolor{mygray-bg}{89.4}}} & \textcolor{blue}{\textbf{\cellcolor{mygray-bg}{23.9}}}  &  \textcolor{blue}{\textbf{\cellcolor{mygray-bg}{0.208}}}  & \textcolor{blue}{\textbf{\cellcolor{mygray-bg}{9.4}}} & \textcolor{blue}{\textbf{\cellcolor{mygray-bg}{15.1}}} & \textcolor{blue}{\textbf{\cellcolor{mygray-bg}{33.3}}} & \textcolor{blue}{\textbf{\cellcolor{mygray-bg}{72.7}}} & \textcolor{blue}{\textbf{\cellcolor{mygray-bg}{19.6}}} & \cellcolor{mygray-bg}{81.4} \\
            \cline{2-16}
                & \textbf{ComPro$^{*}$ (Ours)} & \cellcolor{mygray-bg}{L} & \cellcolor{mygray-bg}{107M} & \cellcolor{mygray-bg}{--} & \cellcolor{mygray-bg}{--} & \cellcolor{mygray-bg}{--} & \cellcolor{mygray-bg}{--} & \cellcolor{mygray-bg}{--} & \cellcolor{mygray-bg}{--}  & \textcolor{red}{\cellcolor{mygray-bg}{9.2}} & \textcolor{red}{\cellcolor{mygray-bg}{15.0}} &  \textcolor{red}{\cellcolor{mygray-bg}{33.3}} &  \textcolor{red}{\cellcolor{mygray-bg}{72.4}} & \cellcolor{mygray-bg}{19.5} &  \textcolor{red}{\cellcolor{mygray-bg}{98.1}} \\
            \hline
        \end{tabular}%
    }
    % \vspace{-1em}
    \caption{Performance of state-of-the-art CIC models and our ComPro framework on COCO Entities and Flickr30K Entities. ``Size" denotes the model size based on different GPT models, \ie, ``B", ``M", and ``L" denote the base, medium, and large models, respectively. ``Params" denotes the number of trainable parameters. The \textcolor{blue}{\textbf{best}} and \textcolor{red}{second best} results are denoted with corresponding formats. $^*$ denotes the models using the probability decay strategy (\cf~appendix) during inference to further control the length of generated captions.}
    % \vspace{-0.5em}
    \label{tab:sota}%
\end{table*}%
\addtolength{\tabcolsep}{2pt}

\textbf{Training Details.} For the region features, we used the bottom-up features~\cite{anderson2018bottom} which are extracted by a Faster R-CNN~\cite{ren2015faster} pre-trained on VG~\cite{sharma2018conceptual}. For the language model, we used the pretrained GPT-2~\cite{radford2019language} with three different model sizes, including GPT-2$_{\text{base}}$, GPT-2$_{\text{medium}}$, and 
GPT-2$_{\text{large}}$. For the prompt generation network, we utilized an eight-layer Transformer encoder with the same hidden size as the GPT-2 (\ie, 768, 1024, and 1280). The content prompt length and structure prompt length ($N, M$) were set to (10, 40) for content control, (40, 40) for structure control, and (10, 40) for combined control, respectively. 
We used the mask attention for all experiments and left the ablation of segment embedding in Sec.~\ref{sec:ablation}.
% The content prompt length and structure prompt length were set to 10 and 40, respectively.
During training, the whole GPT-2 was fixed and we only trained the PGN with a XE loss. 
The batch size was set to 40 and we trained the PGN with Adam optimizer for 10 epochs, and the initial learning rate was set to 2e-5. In inference, we used beam search with size 5.
% In the training stage, the batch size was set to 40 and we trained the PGN with Adam optimizer for 10 epochs, and the initial learning rate was set to 2e-5. In inference, we used beam search with size 5.

\textbf{Signal-specific Settings.} For the content-related control, for both two datasets (\ie, COCO/Flickr30K Entities), we followed the same process as~\cite{cornia2019show} to extract the region sequence for each image and its associated caption. For the structure-related control, for the COCO Entities, we used the same 4-level setting as~\cite{deng2020length}, \ie, the length ranges are $[1,9]$, $[10,14]$, $[15,19]$, and $[20,25]$. For the Flickr30K Entities, according to the distribution of caption length, we used a similar 4-level setting, and the corresponding length ranges are $[1,8]$, $[9,11]$, $[12,14]$, and $[15,25]$. Following the conventions of prior CIC works~\cite{cornia2019show,deng2020length,chen2021human}, we utilized the ground-truth control signals in all the experiments.
% \textcolor{blue}{For the structure-related tense-control, we used the 5-type setting as ~\cite{zhu2021self} for both two datasets, \ie, the tense types are ``no v", ``be + v", ``v-ing", ``v" and ``v-ed".} 

\addtolength{\tabcolsep}{-2.5pt}
\begin{table*}[t]
  \renewcommand\arraystretch{1.05}
  \centering
  \tabcolsep=5pt
  \scalebox{1.1}{
    \begin{tabular}
    {c|c| c |c |c|ccccc|ccc}
    \hline
        % \multirow{2}{*}{Method} &  \multirow{2}{*}{Structure} & \multirow{2}{*}{Content} & \multicolumn{7}{c}{COCO Entities} \\
    % \cline{4-10}           
    \textbf{Datasets} & \textbf{Content} & \textbf{Structure} & \textbf{Tense} & \textbf{Probability Decay} & {B4} & {M} & {R} & {C} & {S} & {NW} & {LP} & {TP}\\
    % \hline
        % \multicolumn{12}{l}{\emph{COCO Entities}} \\
    \hline
       &  \checkmark & &  & &{24.0} & {27.3} & {56.1} & {232.2} & {50.4} & {0.531} & -- & --\\
    
       COCO &  & \checkmark & & &13.9 & {20.8} & 41.9 & 138.5 & 32.0 & -- & 94.7 & --\\
    
      Entities &   \checkmark & \checkmark & & & {24.4} & {27.7} & {56.2} & {235.7} & 50.3 & {0.533} & 88.2 & -- \\
        
        & \checkmark & \checkmark & \checkmark  & & \textbf{28.5} & \textbf{29.2} & \textbf{58.8} & \textbf{269.4} & \textbf{53.0} & \textbf{0.547} & {87.4} & \textbf{90.1}\\
        \cline{2-13}
       &  \checkmark & \checkmark & \checkmark  & \checkmark & 28.0 & 28.8 & 58.4 & 264.7 & 52.6 & 0.542 & \textbf{98.9} & 89.6\\
    % \hline
        % \multicolumn{12}{l}{\emph{Flickr30K Entities}} \\
    \hline
       &  \checkmark & & & &{11.9} & {17.3} & {37.8} & {89.4} & {23.9} & {0.208} & -- & --\\
        
      Flickr30K & & \checkmark & & & {9.4} & {15.1} & {33.3} & {72.7} & {19.6} & -- & 81.4 & --\\

        Entities & \checkmark & \checkmark &  & & {12.1} & {17.3} & {38.2} & {95.8} & {23.8} & {0.211} & 79.4  & -- \\
   
        & \checkmark & \checkmark & \checkmark &  & \textbf{13.5} &  \textbf{17.8} & \textbf{39.7}  &  \textbf{108.6} & \textbf{24.1} &  \textbf{0.229} &  78.9 &  \textbf{82.8}\\
        
        \cline{2-13}
        & \checkmark & \checkmark & \checkmark & \checkmark & 13.3 &  17.5 & 39.4  & 106.9 & 23.8 &  0.223 &  \textbf{97.1} & 81.5 \\
    \hline
    \end{tabular}%
    }
  % \vspace{-1em}
  \caption{{Performance of ComPro with GPT-2$_{\text{large}}$ on combined signals.}}
 % \vspace{-0.5cm}
  \label{tab:combined}%
\end{table*}%
\addtolength{\tabcolsep}{2.5pt}

\subsection{Comparison with State-of-the-Arts}

In this subsection, we evaluated the controllability of our framework on \emph{single} control signals. Specifically, we compared our ComPro with two state-of-the-art CIC models: 1) \textbf{SCT}~\cite{cornia2019show}: it is an auto-regressive region-control CIC model which is built based on UpDn~\cite{anderson2018bottom}. 2) \textbf{LaBERT}~\cite{deng2020length}: it is a non-autoregressive length-control CIC model which is built based on BERT~\cite{devlin2018bert}. Since LaBERT modifies the probability of its stop token during inference to control the caption length, we also apply the same strategy\footnote{More details are left in the appendix.} to our framework, denoted as ComPro$^{*}$. All results are reported in Table~\ref{tab:sota}.

\textbf{Results on COCO Entities.} 
From Table~\ref{tab:sota}, we can have the following observations: 1) For content-related control, all three variants of ComPro achieve significantly better performance than SCT on both accuracy-based metrics and the alignment score NW. Particularly, ComPro with GPT-2$_{\text{small}}$ only has nearly half trainable parameters of SCT (39M vs. 70M), but it can achieve much better performance (\eg, 227.5 vs. 192.3 on CIDEr-D or 0.531 vs. 0.508 on NW). 2) For structure-related control, all three variants of ComPro require fewer trainable parameters than LaBERT. Although ComPro with GPT-2$_{\text{small}}$ and GPT-2$_{\text{medium}}$ achieve comparable performance with LaBERT, ComPro with GPT-2$_{\text{large}}$ can achieve better performance on key accuracy-based metrics (\eg, 13.9 vs. 13.5 on BLEU-4 and 138.5 vs. 136.6 on CIDEr-D). 3) When using the same probability decay strategy as LaBERT, ComPro can also achieve high length precision LP while keeping other metrics almost unchanged.

\textbf{Results on Flickr30K Entities.} 
From Table~\ref{tab:sota}, we can observe: 1) For content-related control, all three variants of ComPro achieve a new state-of-the-art performance on all different metrics. 2) For structure-related control, ComPro with GPT-2$_{\text{large}}$ achieves superior performance than LaBERT with fewer trainable parameters. Since the length range of the control signals on Flickr30K Entities is narrower than COCO Entities, the length precision LP of ComPro on Flickr\-30K Entities is relatively lower. 3) Similarly, when using the same probability decay strategy, ComPro$^{*}$ achieves perfect LP while keeping other metrics high.

\subsection{Combinatorial Ability in ComPro}

\subsubsection{Combination of Multiple Control Types}
To show the combinatorial ability in ComPro, we first evaluated the results on combined controls with the two representative content control and structure control signals.

\addtolength{\tabcolsep}{-1.5pt}
\begin{table*}[t]
  \renewcommand\arraystretch{1.05}
  \centering
  \tabcolsep=5pt
  \scalebox{1.1}{
    \begin{tabular}
    {c |c |c| c|ccccc|cc}
    \hline
        % \multirow{2}{*}{Method} &  \multirow{2}{*}{Structure} & \multirow{2}{*}{Content} & \multicolumn{7}{c}{COCO Entities} \\
    % \cline{4-10}
        \textbf{Dataset} & \textbf{Content} & \textbf{Structure} & \textbf{Probability Decay} & {B4} & {M} & {R} & {C} & {S} & {NW} & {LP} \\
    % \hline
    %     \multicolumn{10}{l}{\emph{COCO Entities}} \\
    \hline
        & \checkmark & &  & {23.6} & {27.1} & {55.6} & {227.5} & 49.9 & {0.519} & -- \\
    
       COCO & & \checkmark & & 13.7 & {20.3} & 41.7 & 134.1 & 31.7 & -- & 92.4 \\
    
        Entities & \checkmark & \checkmark & & \textbf{23.7} & \textbf{27.4} & \textbf{55.6} & \textbf{228.3} & \textbf{49.9} & \textbf{0.526} & 87.3 \\
    \cline{2-11}
       & \checkmark & \checkmark & \checkmark  & 23.5 & 27.2 & 55.4 & 227.7 & 49.6 & 0.522 & \textbf{99.3} \\
    % \hline
    %     \multicolumn{10}{l}{\emph{Flickr30K Entities}} \\
    \hline
       & \checkmark & & & {11.4} & {16.8} & {37.4} & {84.1} & \textbf{23.7} & {0.195} & -- \\
        
       Flickr30K & & \checkmark & &  {9.0} & {14.6} & {32.7} & {67.8} & {19.3} & -- & 80.1 \\

        Entities & \checkmark & \checkmark &  & \textbf{11.4} & \textbf{16.8} & \textbf{37.4} & \textbf{87.3} & {23.2} & \textbf{0.198} & 76.7  \\
    \cline{2-11}
        & \checkmark & \checkmark & \checkmark & 11.3 & 16.6 & 37.2 &  85.8 & 23.0 & 0.195 &  \textbf{98.2} \\
    \hline
    \end{tabular}%
    }
  % \vspace{-1em}
  \caption{Performance of ComPro with GPT-2$_{\text{base}}$ on combined control signals. ``C" and ``S" denote the \emph{content-related} and \emph{structure-related} signals. ``$^*$"(ComPro$^*$) denotes using the probability decay strategy.}
  \label{tab:combined_b}%
\end{table*}%
\addtolength{\tabcolsep}{1.5pt}

\addtolength{\tabcolsep}{-1.5pt}
\begin{table*}[t]
  \renewcommand\arraystretch{1.05}
  \centering
  \tabcolsep=5pt
  \scalebox{1.1}{
    \begin{tabular}
    {c |c| c |c|ccccc|cc}
    \hline
        % \multirow{2}{*}{Method} &  \multirow{2}{*}{Structure} & \multirow{2}{*}{Content} & \multicolumn{7}{c}{COCO Entities} \\
    % \cline{4-10}
       \textbf{Dataset} & Con\textbf{}tent & \textbf{Structure} & \textbf{Probability Decay} & {B4} & {M} & {R} & {C} & {S} & {NW} & {LP} \\
    % \hline
    %     \multicolumn{10}{l}{\emph{COCO Entities}} \\
    \hline
       & \checkmark & &  & {23.5} & {27.2} & {55.6} & {226.1} & \textbf{49.8} & {0.520} & -- \\
    
        COCO & & \checkmark & & 13.6 & {20.5} & 41.6 & 134.8 & 31.4 & -- & 92.5 \\
    
       Entities & \checkmark & \checkmark & & \textbf{23.6} & \textbf{27.3} & \textbf{55.4} & \textbf{228.0} & 49.3 & \textbf{0.521} & 85.5 \\
    \cline{2-11}
        & \checkmark & \checkmark & \checkmark  & 23.4 & 27.2 & 55.2 & 227.5 & 49.2 & 0.518 & \textbf{99.1} \\
    % \hline
    %     \multicolumn{10}{l}{\emph{Flickr30K Entities}} \\
    \hline
       & \checkmark & & & {11.6} & {17.0} & {37.3} & {86.6} & \textbf{23.6} & {0.196} & -- \\
        
      Flickr30K &  & \checkmark & &  {8.9} & {14.7} & {32.6} & {67.9} & {18.9} & -- & 81.8 \\

        Entities & \checkmark & \checkmark &  & \textbf{12.1} & \textbf{17.1} & \textbf{37.8} & \textbf{94.2} & {23.4} & \textbf{0.203} & 77.9  \\
    \cline{2-11}
        & \checkmark & \checkmark & \checkmark & 11.8 & 16.9 & 37.6 & 92.9 & 23.1 & 0.199 &  \textbf{98.0} \\
    \hline
    \end{tabular}%
    }
  % \vspace{-1em}
  \caption{Performance of ComPro with GPT-2$_{\text{medium}}$ on combined control signals. ``C" and ``S" denote the \emph{content-related} and \emph{structure-related} signals. ``$^*$"(ComPro$^*$) denotes using the probability decay strategy.}
  % \vspace{-1em}
  \label{tab:combined_m}%
\end{table*}%
\addtolength{\tabcolsep}{1.5pt}

\begin{figure*}
    \centering
    \includegraphics[width=1.0\linewidth]{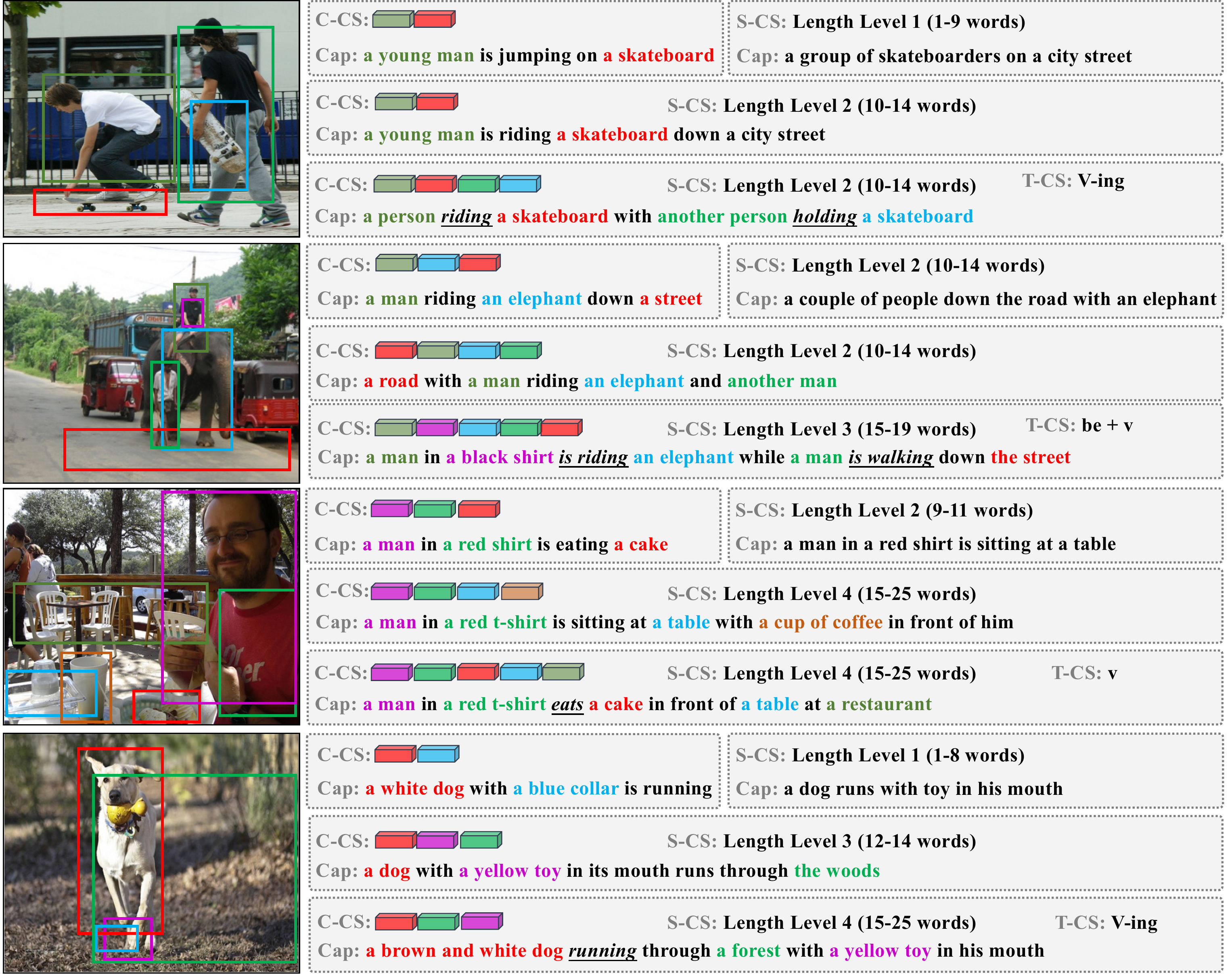}
    % \vspace{-1em}
    \caption{Visualization examples of universal controllable image captioning by ComPro with different control signals. ``C-CS", ``S-CS", ``T-CS", and ``Cap" denote content-related control signal, structure-related control signal, tense control signal, and generated captions, respectively. Different colors show the corresponding control region sets and the associations between generated words and regions.}
    % \vspace{-0.5em}
    \label{fig:visual}
\end{figure*}

\noindent\textbf{Quantitative Results.}
From Table~\ref{tab:combined}, we can observe that: 1) ComPro can efficiently achieve the combinatorial ability on both two control signals with remarkable performance on accuracy-based metrics together with high NW and LP scores. 2) ComPro with combined control can achieve better performance than with single content control, but with lower LP scores than with single length control. As there is a trade-off between generating more semantic information and keeping the caption within specific length ranges.

We reported more results of the ComPro on combined controls with different sizes of GPT-2 models, including ComPro with GPT-2$_{\text{base}}$ in Table~\ref{tab:combined_b} and ComPro with GPT-2$_{\text{medium}}$ in Table~\ref{tab:combined_m}, respectively.
We can observe that: 1) Similar to ComPro with GPT-2$_{\text{large}}$, ComPro with GPT-2$_{\text{base}}$ and GPT-2$_{\text{medium}}$ can also achieve the combinatorial ability on both content and structure control efficiently with outstanding performance on all metrics. 2) While ComPro with combined control achieves the best performance on almost all metrics, ComPro$^*$ can also achieve comparable performance with satisfying LP.

\begin{figure*}[t]
    \centering
    \includegraphics[width=0.7\linewidth]{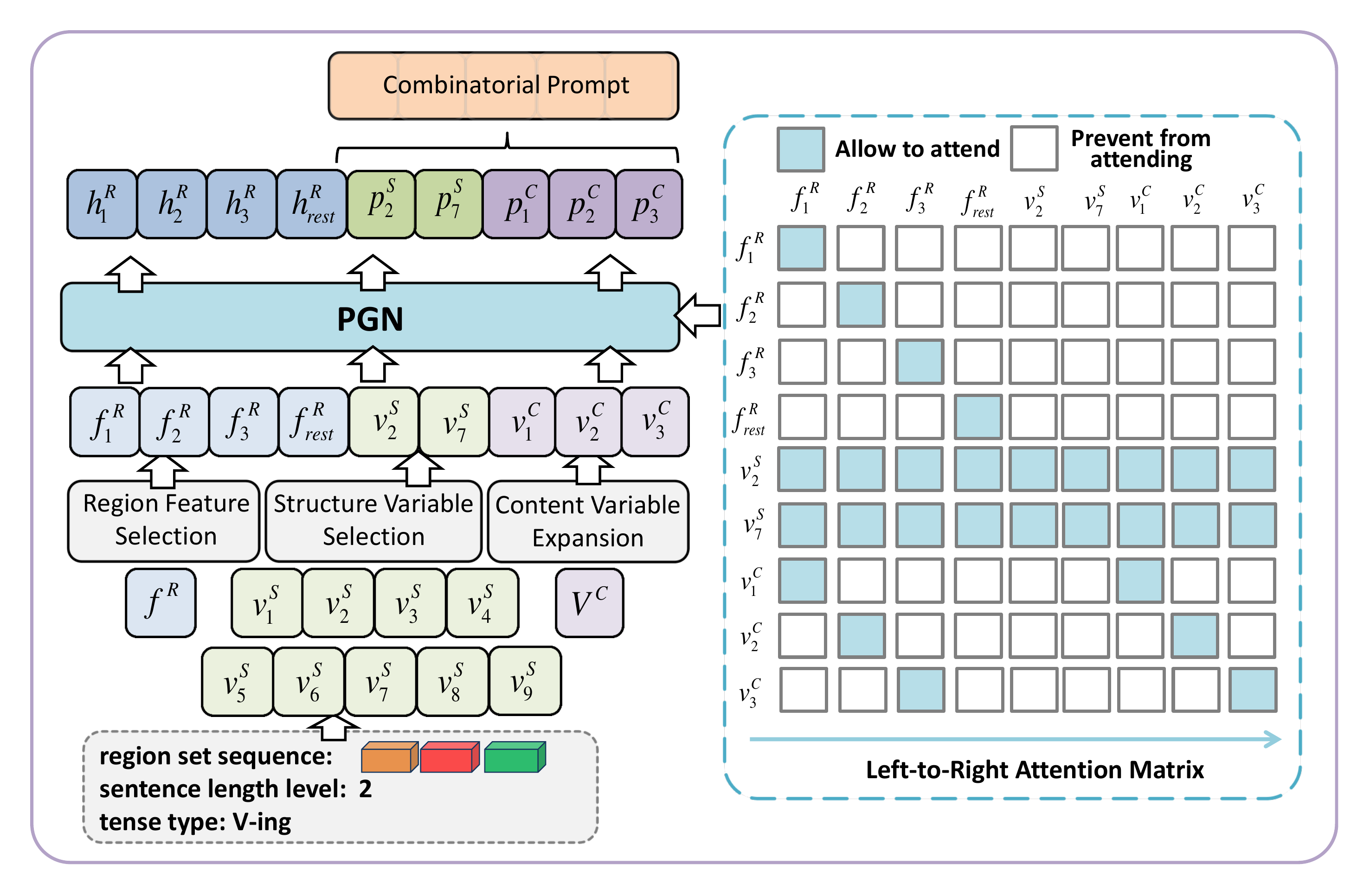}
    % \vspace{-2em}
    \caption{The pipeline of combinatorial prompt generation for extension on tense control. It takes a region set sequence with three objects as the content control signal, the length level-2 (four levels in total), and tense type ``v-ing" (five types in total) as the structure control signals.}
    % \vspace{-1.5em}
    \label{fig:pgn_tense}
\end{figure*}

\noindent\textbf{Qualitative Results.}
We illustrated the generated captions from our ComPro on both single and combined control signals in Fig.~\ref{fig:visual}. We can observe that ComPro can describe each region set by orders (\ie, content control), or generate captions inside specific length ranges (\ie, structure control). Particularly, when facing complex combined controls, our ComPro can generate more details to meet the length constraint and keep the description orders simultaneously.

\subsubsection{Generalization to New Control Signals} \label{sec:4.4.2}

Due to the simple prompt concatenation design philosophy of our universe framework, we then further extended ComPro to more new control signals, \ie, tense control. Strictly speaking, tense control is also a type of structure-related control signal. To avoid confusion, we use ``tense control" to distinguish it from the structure-related control (\ie, length-related) in the whole paper. 

\noindent\textbf{Signal-specific Settings.} As a type of structure-related control signal, the tense control aims to generate captions with given tense types. Following previous works~\cite{zhu2021self}, we used the 5-type setting for both two datasets, \ie, the tense types are ``no v", ``be + v", ``v-ing", ``v" and ``v-ed". And we used the tense precision (\textbf{TP}) to evaluate whether the tense of the generated captions is of desired tense types.

\noindent\textbf{Combinatorial Prompt Generation.} Similar to the generation process of the combined control shown in Sec.~\ref{sec:compo_combine}, ComPro can be easily extended to tense control by generating and concatenating more new sub-prompts. As shown in Fig.~\ref{fig:pgn_tense}, following the same region feature selection and content variable expansion process, we only need to expand more structure variables (\ie, 9 in total, 4 for length level and 5 for tense type) and select the corresponding structure variable for concatenation to generate the combinatorial prompt.

\noindent\textbf{Results.} As shown in Table~\ref{tab:combined}, we can observe that: 1) ComPro can be efficiently extended to new control signals to further improve the performance of the generated captions with desirable control ability for all control signals. 2) ComPro with GPT2$_{\text{large}}$ achieves the best performance on almost all the metrics, and ComPro$^*$ can also achieve comparable performance with satisfying LP.

\subsubsection{Generalization to New Instances} \label{sec:4.4.3}

To further show ComPro has the combinatorial ability to handle various combinations of multiple control instances which are different from those in training samples. We evaluated ComPro with various combinations of region numbers/entities, description orders, length levels, and tense typ\-es.

\noindent\textbf{Results.} As shown in Fig.~\ref{fig:visual}, ComPro successfully meets all constraints simultaneously by generating captions with specific description orders and tense types within given length ranges.

\addtolength{\tabcolsep}{-1.5pt}
\begin{table}[t]
  \centering
  \scalebox{1.1}{
    \begin{tabular}
    {c|c|c|c}
    \hline
        Methods & Model Size & Params & Training Time\\
    \hline
        {SCT~\cite{cornia2019show}} & -- & 70M & 12h\\
    \hline
        {LaBERT~\cite{deng2020length}} & BERT$_{\text{base}}$ & 140M & 18h\\
    \hline
        {\textbf{ComPro}} & GPT-2$_{\text{base}}$ & 39M & 8h\\
        {\textbf{ComPro}} & \quad GPT-2$_{\text{medium}}$ & 69M & 10h\\
        {\textbf{ComPro}} & GPT-2$_{\text{large}}$ & 107M & 12h\\
    \hline
    \end{tabular}%
    }
  % \vspace{-0.5em}
  \caption{Computation cost of ComPro and state-of-the-art CIC models on COCO Entities. ``Params" denotes the number of trainable parameters.}
  % \vspace{-1.0em}
  \label{tab:training_time}%
\end{table}%
\addtolength{\tabcolsep}{1.5pt}

\subsection{Training Efficiency of ComPro}
\label{sec:train_time}
To evaluate the efficiency of our framework, we further report the training time of ComPro and two state-of-the-art CIC models on COCO Entities. Specifically, we report training time in GPU hours, and we used the same A100 GPU with 40G memories for all models. 

From Table~\ref{tab:training_time}, we can observe that: 1) ComPro requires significantly fewer training time and trainable parameters than SCT and LaBERT. 2) ComPro with larger GPT-2 tends to require more trainable parameters thus with more training times for each epoch. However, larger GPT-2 can accelerate the convergence of training, which ends in only a little increase in the whole training time.

\addtolength{\tabcolsep}{-2.5pt}
\begin{table*}[t]
% \small
    \renewcommand\arraystretch{1.05}
    \centering
    \tabcolsep=5pt
  \scalebox{1.0}{
    \begin{tabular}{c|ccccc|c|ccccc|c}
        \hline
        \multirow{2}{*}{Method} & \multicolumn{6}{c|}{{Content-related Control}} & \multicolumn{6}{c}{{Structure-related Control}} \\
        % {Method} & {B4} & {C} & {NW} & {B4} & {C} & {LP}\\
        \cline{2-13}
        & {B4} & {M} & {R} & {C} & {S} & {NW} & {B4} & {M} & {R} & {C} & {S} & {LP} \\
        \hline
        ComPro & \textbf{{23.6}} & \textbf{{27.1}} & \textbf{{55.6}} & \textbf{{227.5}} & \textbf{49.9} & \textbf{{0.519}} & \textbf{13.7} & \textbf{{20.3}} & \textbf{41.7} & \textbf{134.1} & \textbf{31.7} & \textbf{92.4} \\
        \hline
        Type$_1$  & {14.0} & 20.4 & 42.3 & {135.6} & 32.7 & {0.326} & {13.7} & 20.2 & 41.6 & {133.4} & 31.5 & {92.0} \\
        % \hline
        Type$_2$ & {19.0} & 24.7 & 48.9 & {186.3} & 43.9   & {0.402} & {13.6} & 20.1 & 41.6 & {133.6}& 31.6 & {91.2} \\
        \hline
    \end{tabular}%
    }
 % \vspace{-0.5em}
  \caption{Ablations on the mask attention. ``Type$_1$" and ``Type$_2$" denote the baselines removing the mask of all and content variables, respectively.}
  \label{tab:ablation_maskatt}%
    % \vspace{-0.5em}
\end{table*}
\addtolength{\tabcolsep}{2.5pt}

\addtolength{\tabcolsep}{-1.5pt}
\begin{table}[t]
  \renewcommand\arraystretch{1.05}
  \centering
  \scalebox{0.98}{
    \begin{tabular}
    {c | c| c|ccccc|c}
    \hline
        % \multirow{2}{*}{Method} &  \multirow{2}{*}{Structure} & \multirow{2}{*}{Content} & \multicolumn{7}{c}{COCO Entities} \\
    % \cline{4-10}
        Method & MAtt & Seg & {B4} & {M} & {R} & {C} & {S} & {NW}\\
    \hline
        SCT~\cite{cornia2019show} & & & {19.7} & {24.0} & {51.6} & {192.3} & {45.8} & {0.508}\\
    \hline
        & \checkmark& & \textbf{{23.6}} & \textbf{{27.1}} & \textbf{{55.6}} & \textbf{{227.5}} & \textbf{{49.9}} & \textbf{{0.519}}\\
        ComPro & & \checkmark &{23.1} & {26.7} & {54.7} & {220.5} & {48.8} & {0.515}\\
        & \checkmark & \checkmark&{23.4} & {27.1} & {55.5} & {225.2} & {55.5} & {0.519}\\
    \hline
    \end{tabular}%
    }
  % \vspace{-1em}
  \caption{Ablations on the segment embedding. ``MAtt" and ``Seg" denote the mask attention and segment embedding, respectively.}
  \label{tab:ablation_seg}%
\end{table}%
\addtolength{\tabcolsep}{1.5pt}

\subsection{Ablation Studies}
\label{sec:ablation}
In this subsection, we run a set of ablation studies about the ComPro, including the influence of the mask attention mechanisms, segment embedding, prompt length and different visual features. All the experiments are based on GPT-2$_{\text{base}}$ on COCO Entities.

\subsubsection{Mask Attention} To evaluate the influence of our mask attention mechanism, we compared ComPro with two baselines: \textbf{Type$_1$:} Removing the whole mask (\ie, plain self-attention). \textbf{Type$_2$:} Removing the mask of the content variable. 

\textit{\textbf{Results.}} As shown in Table~\ref{tab:ablation_maskatt}, the mask has significant impacts on content control: content variables fail to learn corresponding visual guidance after removing masks. Instead, structure variables designed to learn global information are barely affected.

\subsubsection{Segment Embedding.} To validate the effectiveness of mask attention and segment embedding, we run several ablations with content-related control signals. 

\textit{\textbf{Results.}} From Table~\ref{tab:ablation_seg}, we can observe that: 1) Either ComPro use mask attention or segment embedding significantly outperformed SCT~\cite{cornia2019show}. 2) ComPro with mask attention can achieve the best performance, while the model with only segment embedding or both are not as effective and require more training sources. 
% Thus, such specific segment embedding may also be helpful to achieve controllability, but our mask attention mechanism is much more effective and efficient.

\addtolength{\tabcolsep}{-1.5pt}
\begin{table*}[t]
    \renewcommand\arraystretch{1.05}
    \centering
    \scalebox{1.05}{
        \begin{tabular}
            {c|l|c|ccccc|c|ccccc|c}
            \hline
            \multirow{2}{*}{Datasets} & \multirow{2}{*}{Method} &   \multirow{2}{*}{Visual Features} & \multicolumn{6}{c|}{\textbf{Content-related Control}} & \multicolumn{6}{c}{\textbf{Structure-related Control}} \\
            \cline{4-15}
            & & & {B4} & {M} & {R} & {C} & {S} & {NW} & {B4} & {M} & {R} & {C} & {S} & {LP} \\
            \hline
   \multirow{3}{*}{COCO Entities} & \textbf{ComPro} & BUTD~\cite{anderson2018bottom} & {23.6} & 27.1 & 55.6 & {227.5} & {49.9} & 0.519 & 13.7 & 20.3 & 41.7 & 134.1 & 31.7  & 92.4  \\
    & \textbf{ComPro} & CLIP\_ViT & 20.2 & 24.9 & 51.5 & 193.9 & 44.3 & 0.458 & 13.3 & 20.4 & 41.7 & 132.7 & 31.6 & 92.7  \\ 
    & \textbf{ComPro} & CLIP\_RS & 20.5 & 24.8 & 51.6 & 194.1 & 44.0 & 0.465 & 13.5 & 20.4 & 41.7 & 134.5 & 31.7 & 92.8  \\ 
    \hline
        \end{tabular}%
    }
    % \vspace{-0.5em}
    \caption{Performance of our ComPro framework with different visual features. ``BUTD" denotes the bottom-up features. ``CLIP\_ViT" and ``CLIP\_RN" denote the CLIP features extracted by ViT and ResNet visual encoders of CLIP, respectively.}
    \label{tab:feat}%
\end{table*}%
\addtolength{\tabcolsep}{1.5pt}

\addtolength{\tabcolsep}{-1.5pt}
\begin{table}[t]
    \renewcommand\arraystretch{1.05}
    % \vspace{-1em}
    \begin{center}
    \scalebox{1.05}{
        \begin{tabular}{c|c|ccccc|c}
        \hline 
        {Dataset} & {SP Len} & {B-4} & {M} & {R} & {C} & {S} & {LP}\\
        \hline
        & 1 & 12.4 & 19.3 & 40.1 & 123.8 & 30.8 & 76.9 \\
        & 5 & 13.2 & 20.1 & 41.3 & 130.5 & 31.1 & 88.8 \\
        COCO & 10 & 13.2 & 20.2 & 41.4 & 130.3 & 31.2 & 89.9 \\
        Entities & 20 & 13.3 & \textbf{20.3} & 41.6 & 132.8 & \textbf{31.7} & 91.3 \\
        & 40 & \textbf{13.7} & 20.3 & \textbf{41.7} & \textbf{134.1} & 31.7 & \textbf{92.4} \\
        & 80 & 13.5 & 20.3 & 41.7 & 133.9 & 31.6 & 91.5 \\
        \hline
        \end{tabular}%
    }
    \end{center}
    % \vspace{-1.5em}
    \caption{Ablation studies on the structure control performance of ComPro with different structure prompt (SP) lengths. }
    \label{tab:ablation_splength}%
\end{table}%
\addtolength{\tabcolsep}{1.5pt}

\subsubsection{Length of Structure Prompt.} We first evaluated the performance of ComPro on structure control by setting different structure prompt lengths $M \in \{1, 5, 10, 20, 40, 80\}$. To explore the influence of $M$, content prompt length $N$ was set the same as $M$ in this study. 

\textit{\textbf{Results.}} From Table~\ref{tab:ablation_splength}, we can observe that: 1) The performance of ComPro keeps improving when we increase the prompt length from 1 to 20. Then, the metrics reach the best scores and keep unchanged or even slightly drop with longer prompt length. For example, CIDEr-D reaches the best score 134.1 with $M=40$ and drops to 133.9 with $M=80$. Since most metrics reach their best scores with $M=40$, we set $M$ to 40 in our experiments.

\subsubsection{Length of Content Prompt.} We then evaluated the performance of ComPro on content control by setting different content prompt lengths $N \in \{1, 5, 8, 10, 15, 20\}$ to explore the influence of $N$.

\textit{\textbf{Results.}} From Table~\ref{tab:ablation_cplength}, we have several observations: 1) The performance of ComPro keeps improving when we increase the prompt length from 1 to 10, and all metrics reach the best scores with prompt length $N=10$. 2) Together with the results in Table~\ref{tab:ablation_splength}, prompts with extremely small lengths tend to generate inferior performance. Meanwhile, prompts with much longer lengths lead to performance saturation. Thus, considering the trade-off between model performance and training time, we set $N$ to 10.

\addtolength{\tabcolsep}{-1.5pt}
\begin{table}[t]
  \renewcommand\arraystretch{1.05}
  % \vspace{-1em}
  \begin{center}
  \scalebox{1.05}{
    \begin{tabular}{c|c|ccccc|c}
    \hline 
    Dataset & {CP Len} & {B-4} & {M} & {R} & {C} & {S} & {NW}\\
    \hline
    & 1 & 22.6 & 26.5 & 54.4 & 217.2 & 48.1 & 0.499 \\
    & 5 & 22.8 & 26.8 & 55.1 & 220.8 & 49.3 & 0.515 \\
    COCO & 8 & 23.0 & 26.9 & 55.3 & 221.6 & 49.3 & 0.517 \\
    Entities & 10 & \textbf{23.6} & \textbf{27.1} & \textbf{55.6 }& \textbf{227.5} & \textbf{49.9} & \textbf{0.519} \\
    & 15 & 23.2 & 26.9 & 55.0 & 222.7 & 49.2 & 0.514 \\
    & 20 & 23.2 & 26.9 & 55.4 & 223.9 & 49.7 & 0.513 \\
    \hline
    \end{tabular}%
  }
  \end{center}
  % \vspace{-1.5em}
  \caption{Ablation studies on the content control performance of ComPro with different content prompt (CP) lengths.}
  \label{tab:ablation_cplength}%
\end{table}%
\addtolength{\tabcolsep}{1.5pt}

% \textbf{Prompt Length.} We evaluated the performance of ComPro on structure control and content control by setting different structure prompt lengths $M \in \{1, 10, 40, 80\}$ and content prompt lengths $N \in \{1, 5, 10, 20\}$, respectively. From Table~\ref{tab:ablation_plength}, we can observe that: The performance of ComPro keeps improving when we increase the prompt length. Then, the metrics reach the best scores and even slightly drop with longer prompt length. 2) Prompts with extremely small lengths tend to generate inferior performance. Meanwhile, prompts with much longer lengths lead to performance saturation. Thus, considering the trade-off between model performance and training time, we set $M$ to 40 and $N$ to 10 in our experiments.

\subsubsection{Visual Features.} We run a set of ablation studies about the influence of different visual features on ComPro, including the bottom-up features~\cite{anderson2018bottom} and the CLIP~\cite{radford2021learning} features with ResNet or ViT~\cite{dosovitskiy2020image} visual encoders. Specifically, we feed the region boxes from the bottom-up features~\cite{anderson2018bottom} into the CLIP visual encoders to extract the CLIP features for each region. 

\textit{\textbf{Results.}} From Table~\ref{tab:feat}, we can observe that: 1) For structure-related control, ComPro trained with different features achieve comparable performance on all metrics. 2) For conte\-nt-related control, ComPro trained with CLIP features still achieves comparable performance with existing state-of-the-art CIC models, but ComPro trained with bottom-up features can achieve significantly better performance. 3) Since ComPro mainly focuses on the global visual information for structure-related control, the features that are trained to match vision and language (\ie, CLIP features) can provide enough global information for the image. However, the goal of content-related control is to learn the visual relationships within each region set with the same object class, thus the features based on object detection (\ie, bottom-up features) can naturally achieve better performance.

\subsection{User Studies}

\noindent\textbf{User Study Interface.} 
We conducted user studies on content-related control to further evaluate the effectiveness of ComPro. Specifically, we invite 5 experts and give them an image, content control signals (\ie, control regions and orders), and two captions that describe it. They are asked to choose which one better matches the image and control signals in terms of accuracy and controllability and the user interface for this is shown in Appendix. 

\noindent\textbf{User Study Settings.} We randomly select 100 trials from the test split of COCO Entities and ask the experts to give their judgments, each trial contains one image, control signals, and two captions generated by SCT~\cite{cornia2019show} and ComPro, respectively. The caption which got more than 3 votes is regarded as human judgment. Meanwhile, we also calculated the agreement counts between human judgments and other metrics (\ie, whether humans and metrics give higher scores to the same caption).

\addtolength{\tabcolsep}{-2.5pt}
\begin{table}[t]
    \renewcommand\arraystretch{1.05}
    \centering
    \scalebox{1.1}{
	\begin{tabular}{cccc}
	\hline
        Model & C~\cite{vedantam2015cider} & NW~\cite{cornia2019show} & Human Judgements\\
        \hline
        SCT~\cite{cornia2019show} & 168.3 & 0.362 & 14\\
        \textbf{ComPro} & 252.3 & 0.786 & 86\\
        \hline
        \hline
        {Agreements} & {80} & {84} & {-}\\
        \hline
	\end{tabular}
    }
    % \vspace{-0.5em}
     \caption{\textbf{Results of the user studies}. }
    % \vspace{-1.0em}
    \label{tab:users}
\end{table}
\addtolength{\tabcolsep}{3pt}

\noindent\textbf{Results.} As shown in Table~\ref{tab:users}, ComPro achieves better performance on all metrics as well as human judgments (HJ). Meanwhile, both CIDEr-D~\cite{vedantam2015cider} and NW~\cite{cornia2019show} achieve high agreement with humans, \ie, whether humans and metrics give higher scores to the same captions.

% Since we evaluated the controllability of CIC models together with accuracy-based metrics and the NW~\cite{cornia2019show} score globally, we conducted user studies\footref{footnote:appendix} on the content-related control to further validate the effectiveness of these metrics. We randomly selected 100 trials from the test set, each trial contains one image, control signals, two captions generated by SCT~\cite{cornia2019show} and ComPro, respectively. 5 experts are asked to choose the better caption in terms of accuracy and controllability. We then calculated the agreements between human judgments and different metrics (\ie,whether humans and metrics give higher scores to a same caption). As the results shown in Table~\ref{tab:users}, NW~\cite{cornia2019show} achieves high agreement with human judgements similar to the CIDEr-D~\cite{vedantam2015cider}.
% both CIDEr-D~\cite{vedantam2015cider} and NW~\cite{cornia2019show} achieves high agreement with human judgements.\footref{footnote:appendix}

\section{Conclusion and Future Work}
\label{sec:conclusion}
In this paper, we pointed out the importance of the combinatorial ability of CIC models and argued that all existing CIC models have overlooked it. To this end, we proposed a new prompt-based framework ComPro for universal CIC. ComPro can not only achieve state-of-the-art captioning performance under different single control signals, but also be easily extended to more complex combined control signals. We have validated the effectiveness of ComPro through extensive ablations on two challenging benchmarks: COCO Entities and Flickr30K Entities. Moving forward, we are going to: 1) extend our framework into other controllable text generation tasks, \eg, controllable video captioning; 2) design better strategies to utilize these generated prompts for different pretrain language or VL models.

\noindent
\\
\small
\noindent\textbf{Data Availability}
All experiments are conducted on publicly available datasets; see the references cited.

% \section*{Appendix}
% \appendix

\bibliographystyle{spbasic}      % basic style, author-year citations
\bibliography{egbib.bib}

\begin{thebibliography}{55}
\providecommand{\natexlab}[1]{#1}
\providecommand{\url}[1]{{#1}}
\providecommand{\urlprefix}{URL }
\expandafter\ifx\csname urlstyle\endcsname\relax
  \providecommand{\doi}[1]{DOI~\discretionary{}{}{}#1}\else
  \providecommand{\doi}{DOI~\discretionary{}{}{}\begingroup
  \urlstyle{rm}\Url}\fi
\providecommand{\eprint}[2][]{\url{#2}}

\bibitem[{Anderson et~al.(2016)Anderson, Fernando, Johnson, and
  Gould}]{anderson2016spice}
Anderson P, Fernando B, Johnson M, Gould S (2016) Spice: Semantic propositional
  image caption evaluation. In: ECCV, pp 382--398

\bibitem[{Anderson et~al.(2018)Anderson, He, Buehler, Teney, Johnson, Gould,
  and Zhang}]{anderson2018bottom}
Anderson P, He X, Buehler C, Teney D, Johnson M, Gould S, Zhang L (2018)
  Bottom-up and top-down attention for image captioning and visual question
  answering. In: CVPR, pp 6077--6086

\bibitem[{Banerjee and Lavie(2005)}]{banerjee2005meteor}
Banerjee S, Lavie A (2005) Meteor: An automatic metric for mt evaluation with
  improved correlation with human judgments. In: ACL workshop, pp 65--72

\bibitem[{Brown et~al.(2020)Brown, Mann, Ryder, Subbiah, Kaplan, Dhariwal,
  Neelakantan, Shyam, Sastry, Askell et~al.}]{brown2020language}
Brown T, Mann B, Ryder N, Subbiah M, Kaplan JD, Dhariwal P, Neelakantan A,
  Shyam P, Sastry G, Askell A, et~al. (2020) Language models are few-shot
  learners. NeurIPS 33:1877--1901

\bibitem[{Chen et~al.(2022)Chen, Guo, Yi, Li, and
  Elhoseiny}]{chen2022visualgpt}
Chen J, Guo H, Yi K, Li B, Elhoseiny M (2022) Visualgpt: Data-efficient
  adaptation of pretrained language models for image captioning. In: CVPR, pp
  18030--18040

\bibitem[{Chen et~al.(2017)Chen, Zhang, Xiao, Nie, Shao, Liu, and
  Chua}]{chen2017sca}
Chen L, Zhang H, Xiao J, Nie L, Shao J, Liu W, Chua TS (2017) Sca-cnn: Spatial
  and channel-wise attention in convolutional networks for image captioning.
  In: CVPR, pp 5659--5667

\bibitem[{Chen et~al.(2021)Chen, Jiang, Xiao, and Liu}]{chen2021human}
Chen L, Jiang Z, Xiao J, Liu W (2021) Human-like controllable image captioning
  with verb-specific semantic roles. In: CVPR, pp 16846--16856

\bibitem[{Chen et~al.(2020)Chen, Jin, Wang, and Wu}]{chen2020say}
Chen S, Jin Q, Wang P, Wu Q (2020) Say as you wish: Fine-grained control of
  image caption generation with abstract scene graphs. In: CVPR, pp 9962--9971

\bibitem[{Chen et~al.(2018)Chen, Zhang, You, Fang, Wang, Jin, and
  Luo}]{chen2018factual}
Chen T, Zhang Z, You Q, Fang C, Wang Z, Jin H, Luo J (2018)
  ``factual''or``emotional'': Stylized image captioning with adaptive learning
  and attention. In: ECCV, pp 519--535

\bibitem[{Cornia et~al.(2019)Cornia, Baraldi, and Cucchiara}]{cornia2019show}
Cornia M, Baraldi L, Cucchiara R (2019) Show, control and tell: A framework for
  generating controllable and grounded captions. In: CVPR, pp 8307--8316

\bibitem[{Dai et~al.(2017)Dai, Fidler, Urtasun, and Lin}]{dai2017towards}
Dai B, Fidler S, Urtasun R, Lin D (2017) Towards diverse and natural image
  descriptions via a conditional gan. In: ICCV, pp 2970--2979

\bibitem[{Deng et~al.(2020)Deng, Ding, Tan, and Wu}]{deng2020length}
Deng C, Ding N, Tan M, Wu Q (2020) Length-controllable image captioning. In:
  ECCV, pp 712--729

\bibitem[{Deshpande et~al.(2019)Deshpande, Aneja, Wang, Schwing, and
  Forsyth}]{deshpande2019fast}
Deshpande A, Aneja J, Wang L, Schwing AG, Forsyth D (2019) Fast, diverse and
  accurate image captioning guided by part-of-speech. In: CVPR, pp 10695--10704

\bibitem[{Devlin et~al.(2018)Devlin, Chang, Lee, and
  Toutanova}]{devlin2018bert}
Devlin J, Chang MW, Lee K, Toutanova K (2018) Bert: Pre-training of deep
  bidirectional transformers for language understanding. arXiv

\bibitem[{Dosovitskiy et~al.(2020)Dosovitskiy, Beyer, Kolesnikov, Weissenborn,
  Zhai, Unterthiner, Dehghani, Minderer, Heigold, Gelly
  et~al.}]{dosovitskiy2020image}
Dosovitskiy A, Beyer L, Kolesnikov A, Weissenborn D, Zhai X, Unterthiner T,
  Dehghani M, Minderer M, Heigold G, Gelly S, et~al. (2020) An image is worth
  16x16 words: Transformers for image recognition at scale. arXiv

\bibitem[{Gan et~al.(2017)Gan, Gan, He, Gao, and Deng}]{gan2017stylenet}
Gan C, Gan Z, He X, Gao J, Deng L (2017) Stylenet: Generating attractive visual
  captions with styles. In: CVPR, pp 3137--3146

\bibitem[{Han et~al.(2021)Han, Zhao, Ding, Liu, and Sun}]{han2021ptr}
Han X, Zhao W, Ding N, Liu Z, Sun M (2021) Ptr: Prompt tuning with rules for
  text classification. arXiv

\bibitem[{Karpathy and Fei-Fei(2015)}]{karpathy2015deep}
Karpathy A, Fei-Fei L (2015) Deep visual-semantic alignments for generating
  image descriptions. In: CVPR, pp 3128--3137

\bibitem[{Keskar et~al.(2019)Keskar, McCann, Varshney, Xiong, and
  Socher}]{keskar2019ctrl}
Keskar NS, McCann B, Varshney LR, Xiong C, Socher R (2019) Ctrl: A conditional
  transformer language model for controllable generation. arXiv

\bibitem[{Kikuchi et~al.(2016)Kikuchi, Neubig, Sasano, Takamura, and
  Okumura}]{kikuchi2016controlling}
Kikuchi Y, Neubig G, Sasano R, Takamura H, Okumura M (2016) Controlling output
  length in neural encoder-decoders. arXiv

\bibitem[{Lester et~al.(2021)Lester, Al-Rfou, and Constant}]{lester2021power}
Lester B, Al-Rfou R, Constant N (2021) The power of scale for
  parameter-efficient prompt tuning. arXiv

\bibitem[{Li et~al.(2022)Li, Li, Xiong, and Hoi}]{li2022blip}
Li J, Li D, Xiong C, Hoi S (2022) Blip: Bootstrapping language-image
  pre-training for unified vision-language understanding and generation. arXiv

\bibitem[{Li and Liang(2021)}]{li2021prefix}
Li XL, Liang P (2021) Prefix-tuning: Optimizing continuous prompts for
  generation. arXiv

\bibitem[{Lin(2004)}]{lin2004rouge}
Lin CY (2004) Rouge: A package for automatic evaluation of summaries. In: ACL
  workshop, pp 74--81

\bibitem[{Lin et~al.(2014)Lin, Maire, Belongie, Hays, Perona, Ramanan,
  Doll{\'a}r, and Zitnick}]{lin2014microsoft}
Lin TY, Maire M, Belongie S, Hays J, Perona P, Ramanan D, Doll{\'a}r P, Zitnick
  CL (2014) Microsoft coco: Common objects in context. In: ECCV, pp 740--755

\bibitem[{Lindh et~al.(2020)Lindh, Ross, and Kelleher}]{lindh2020language}
Lindh A, Ross RJ, Kelleher JD (2020) Language-driven region pointer advancement
  for controllable image captioning. arXiv

\bibitem[{Liu et~al.(2021)Liu, Yuan, Fu, Jiang, Hayashi, and
  Neubig}]{liu2021pre}
Liu P, Yuan W, Fu J, Jiang Z, Hayashi H, Neubig G (2021) Pre-train, prompt, and
  predict: A systematic survey of prompting methods in natural language
  processing. arXiv

\bibitem[{Luo et~al.(2022{\natexlab{a}})Luo, Xi, Zhang, and Ma}]{luo2022tuning}
Luo Z, Xi Y, Zhang R, Ma J (2022{\natexlab{a}}) I-tuning: Tuning language
  models with image for caption generation. arXiv

\bibitem[{Luo et~al.(2022{\natexlab{b}})Luo, Xi, Zhang, and Ma}]{luo2022vc}
Luo Z, Xi Y, Zhang R, Ma J (2022{\natexlab{b}}) Vc-gpt: Visual conditioned gpt
  for end-to-end generative vision-and-language pre-training. arXiv

\bibitem[{Mathews et~al.(2016)Mathews, Xie, and He}]{mathews2016senticap}
Mathews A, Xie L, He X (2016) Senticap: Generating image descriptions with
  sentiments. In: AAAI, vol~30

\bibitem[{Mathews et~al.(2018)Mathews, Xie, and He}]{mathews2018semstyle}
Mathews A, Xie L, He X (2018) Semstyle: Learning to generate stylised image
  captions using unaligned text. In: CVPR, pp 8591--8600

\bibitem[{Mokady et~al.(2021)Mokady, Hertz, and Bermano}]{mokady2021clipcap}
Mokady R, Hertz A, Bermano AH (2021) Clipcap: Clip prefix for image captioning.
  arXiv

\bibitem[{Needleman and Wunsch(1970)}]{needleman1970general}
Needleman SB, Wunsch CD (1970) A general method applicable to the search for
  similarities in the amino acid sequence of two proteins. Journal of molecular
  biology 48(3):443--453

\bibitem[{Papineni et~al.(2002)Papineni, Roukos, Ward, and
  Zhu}]{papineni2002bleu}
Papineni K, Roukos S, Ward T, Zhu WJ (2002) Bleu: a method for automatic
  evaluation of machine translation. In: ACL, pp 311--318

\bibitem[{Plummer et~al.(2015)Plummer, Wang, Cervantes, Caicedo, Hockenmaier,
  and Lazebnik}]{plummer2015flickr30k}
Plummer BA, Wang L, Cervantes CM, Caicedo JC, Hockenmaier J, Lazebnik S (2015)
  Flickr30k entities: Collecting region-to-phrase correspondences for richer
  image-to-sentence models. In: ICCV, pp 2641--2649

\bibitem[{Radford et~al.(2018)Radford, Narasimhan, Salimans, Sutskever
  et~al.}]{radford2018improving}
Radford A, Narasimhan K, Salimans T, Sutskever I, et~al. (2018) Improving
  language understanding by generative pre-training. OpenAI

\bibitem[{Radford et~al.(2019)Radford, Wu, Child, Luan, Amodei, Sutskever
  et~al.}]{radford2019language}
Radford A, Wu J, Child R, Luan D, Amodei D, Sutskever I, et~al. (2019) Language
  models are unsupervised multitask learners. OpenAI blog 1(8):9

\bibitem[{Radford et~al.(2021)Radford, Kim, Hallacy, Ramesh, Goh, Agarwal,
  Sastry, Askell, Mishkin, Clark et~al.}]{radford2021learning}
Radford A, Kim JW, Hallacy C, Ramesh A, Goh G, Agarwal S, Sastry G, Askell A,
  Mishkin P, Clark J, et~al. (2021) Learning transferable visual models from
  natural language supervision. In: ICML, pp 8748--8763

\bibitem[{Ren et~al.(2015)Ren, He, Girshick, and Sun}]{ren2015faster}
Ren S, He K, Girshick R, Sun J (2015) Faster r-cnn: Towards real-time object
  detection with region proposal networks. NeurIPS

\bibitem[{Schick and Sch{\"u}tze(2020)}]{schick2020exploiting}
Schick T, Sch{\"u}tze H (2020) Exploiting cloze questions for few shot text
  classification and natural language inference. arXiv

\bibitem[{Sharma et~al.(2018)Sharma, Ding, Goodman, and
  Soricut}]{sharma2018conceptual}
Sharma P, Ding N, Goodman S, Soricut R (2018) Conceptual captions: A cleaned,
  hypernymed, image alt-text dataset for automatic image captioning. In: ACL,
  pp 2556--2565

\bibitem[{Shuster et~al.(2019)Shuster, Humeau, Hu, Bordes, and
  Weston}]{shuster2019engaging}
Shuster K, Humeau S, Hu H, Bordes A, Weston J (2019) Engaging image captioning
  via personality. In: CVPR, pp 12516--12526

\bibitem[{Su et~al.(2022)Su, Lan, Liu, Liu, Yogatama, Wang, Kong, and
  Collier}]{su2022language}
Su Y, Lan T, Liu Y, Liu F, Yogatama D, Wang Y, Kong L, Collier N (2022)
  Language models can see: Plugging visual controls in text generation. arXiv

\bibitem[{Sun et~al.(2021)Sun, Wang, Feng, Ding, Pang, Shang, Liu, Chen, Zhao,
  Lu et~al.}]{sun2021ernie}
Sun Y, Wang S, Feng S, Ding S, Pang C, Shang J, Liu J, Chen X, Zhao Y, Lu Y,
  et~al. (2021) Ernie 3.0: Large-scale knowledge enhanced pre-training for
  language understanding and generation. arXiv

\bibitem[{Tang et~al.(2022)Tang, Li, Zhao, and Wen}]{tang2022context}
Tang T, Li J, Zhao WX, Wen JR (2022) Context-tuning: Learning contextualized
  prompts for natural language generation. arXiv

\bibitem[{Vaswani et~al.(2017)Vaswani, Shazeer, Parmar, Uszkoreit, Jones,
  Gomez, Kaiser, and Polosukhin}]{vaswani2017attention}
Vaswani A, Shazeer N, Parmar N, Uszkoreit J, Jones L, Gomez AN, Kaiser {\L},
  Polosukhin I (2017) Attention is all you need. NeurIPS 30

\bibitem[{Vedantam et~al.(2015)Vedantam, Lawrence~Zitnick, and
  Parikh}]{vedantam2015cider}
Vedantam R, Lawrence~Zitnick C, Parikh D (2015) Cider: Consensus-based image
  description evaluation. In: CVPR, pp 4566--4575

\bibitem[{Vinyals et~al.(2015)Vinyals, Toshev, Bengio, and
  Erhan}]{vinyals2015show}
Vinyals O, Toshev A, Bengio S, Erhan D (2015) Show and tell: A neural image
  caption generator. In: CVPR, pp 3156--3164

\bibitem[{Wang et~al.(2020)Wang, Xu, Wang, and Chan}]{wang2020compare}
Wang J, Xu W, Wang Q, Chan AB (2020) Compare and reweight: Distinctive image
  captioning using similar images sets. In: ECCV, pp 370--386

\bibitem[{Xu et~al.(2015)Xu, Ba, Kiros, Cho, Courville, Salakhudinov, Zemel,
  and Bengio}]{xu2015show}
Xu K, Ba J, Kiros R, Cho K, Courville A, Salakhudinov R, Zemel R, Bengio Y
  (2015) Show, attend and tell: Neural image caption generation with visual
  attention. In: ICML, pp 2048--2057

\bibitem[{Yang et~al.(2022)Yang, Liu, Lei, Yang, Xue, Chen, and
  Xie}]{yang2022tailor}
Yang K, Liu D, Lei W, Yang B, Xue M, Chen B, Xie J (2022) Tailor: A
  prompt-based approach to attribute-based controlled text generation. arXiv

\bibitem[{Young et~al.(2014)Young, Lai, Hodosh, and
  Hockenmaier}]{young2014image}
Young P, Lai A, Hodosh M, Hockenmaier J (2014) From image descriptions to
  visual denotations: New similarity metrics for semantic inference over event
  descriptions. TACL 2:67--78

\bibitem[{Zheng et~al.(2019)Zheng, Li, and Wang}]{zheng2019intention}
Zheng Y, Li Y, Wang S (2019) Intention oriented image captions with guiding
  objects. In: CVPR, pp 8395--8404

\bibitem[{Zhong et~al.(2020)Zhong, Wang, Chen, Yu, and
  Li}]{zhong2020comprehensive}
Zhong Y, Wang L, Chen J, Yu D, Li Y (2020) Comprehensive image captioning via
  scene graph decomposition. In: ECCV, pp 211--229

\bibitem[{Zhu et~al.(2021)Zhu, Wang, and Qu}]{zhu2021self}
Zhu Z, Wang T, Qu H (2021) Self-annotated training for controllable image
  captioning. arXiv

\end{thebibliography}


% % Non-BibTeX users please use
% \begin{thebibliography}{}
% %
% % and use \bibitem to create references. Consult the Instructions
% % for authors for reference list style.
% % Format for Journal Reference
% Author, Article title, Journal, Volume, page numbers (year)
% Author, Book title, page numbers. Publisher, place (year)
% \end{thebibliography}

\appendix

\section*{Appendix}
\appendix

\begin{figure*}[t]
    \centering
    \includegraphics[width=0.8\linewidth]{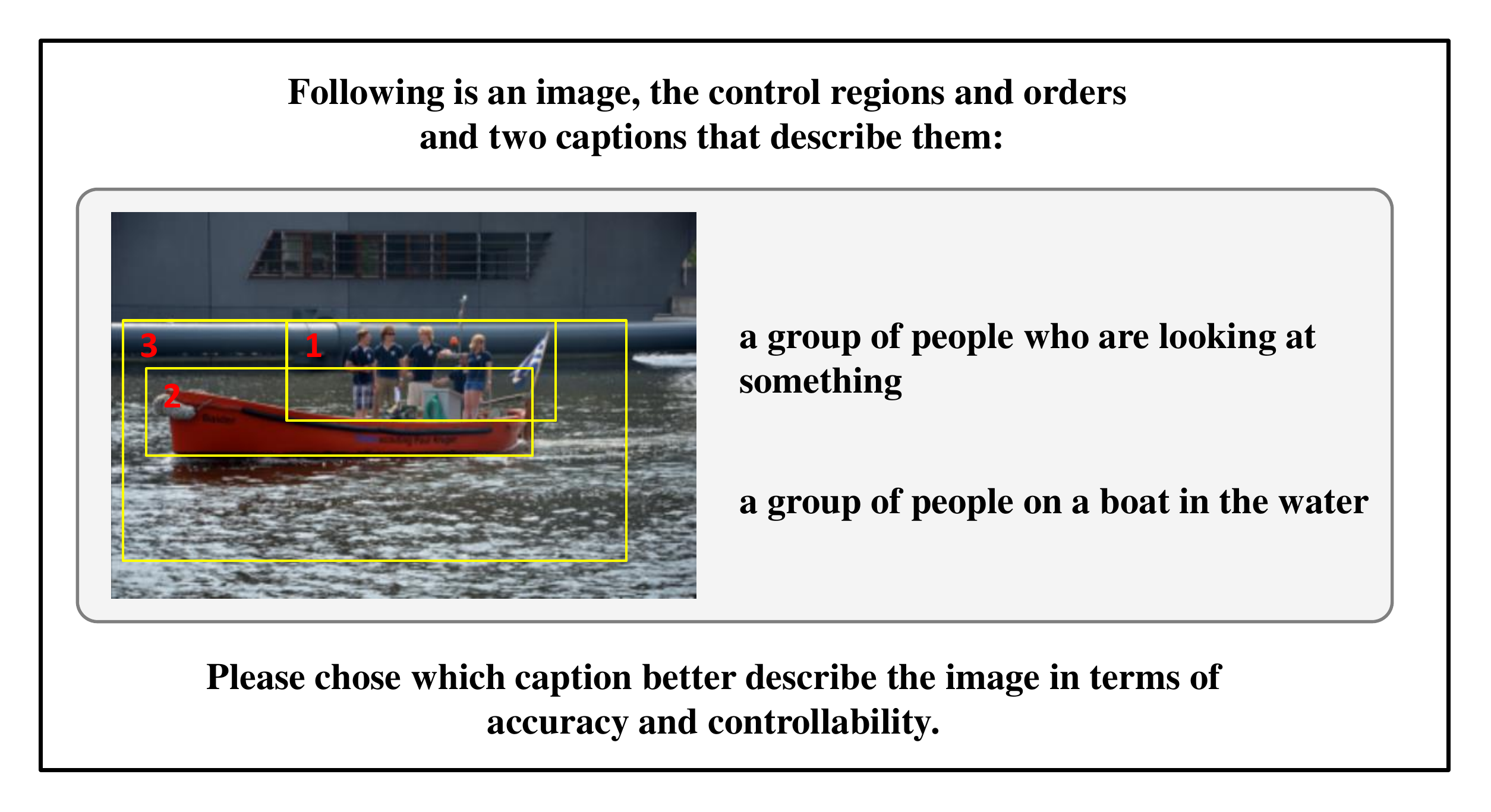}
    % \vspace{-2em}
    \caption{We display an image, the control regions and orders, and two captions generated by SCT~\cite{cornia2019show}and our proposed ComPro, respectively. The
users are asked to choose which caption better describe the image in terms of accuracy and controllability.}
    \label{fig:us}
\end{figure*}

\section{Probability Decay Strategy}

For the length-control, we followed the probability decay strategy of LaBERT~\cite{deng2020length} to further control the caption length during inference.
Specifically, for each sentence length level with range $[L_{low}, L_{high}]$, in order to encourage the model to predict longer captions, we exponentially decay the probability of the stop token $[\mathtt{EOS}]$ by a factor $\gamma$ for predictions after $L_{low}$:
\begin{equation}
\begin{gathered}
p(y_i=[\mathtt{EOS}]) \leftarrow \gamma^{L_{high} - i}p(y_i=[\mathtt{EOS}]), \\
\forall i \in [L_{low},L_{high}]
\end{gathered}
\end{equation}
where $y_i$ is the $i$-th token of the whole sentence. Specifically, $\gamma$ is set to 0.90 and 0.85 for COCO Entities and Flickr30K Entities, respectively. 

Meanwhile, since LaBERT only feeds $L_{high}$ consecutive $[\mathtt{MASK}]$ tokens for prediction, it can ensure the caption length will not exceed the upper bound of length range, \ie, $L_{high}$. To follow such a ``strong" control process, we further limit the caption length by early stopping the prediction when the caption length reaches $L_{high}$.

\section{More Details for User Study}

The user interface is shown in Fig.~\ref{fig:us}.

\end{document}